\documentclass[10pt,twocolumn,letterpaper]{article}

\usepackage{iccv}
\usepackage{times}
\usepackage{epsfig}
\usepackage{graphicx}
\usepackage{amsmath}
\usepackage{amssymb}
\usepackage{enumitem}
\usepackage{arydshln}
\usepackage{xfrac}

\usepackage[pagebackref=true,breaklinks=true,letterpaper=true,colorlinks,bookmarks=false]{hyperref}
\usepackage{color,xcolor}
\usepackage{epsfig}
\usepackage{graphicx}
\usepackage{times}

\usepackage{comment}

\usepackage{pifont}

\usepackage{microtype}
\usepackage[font=small]{caption}
\usepackage{arydshln}
\usepackage{tabularx}
\usepackage{adjustbox}
\usepackage{array}
\usepackage{booktabs}
\usepackage{colortbl}
\usepackage{float,wrapfig}
\usepackage{hhline}
\usepackage{multirow}
\usepackage{subcaption} %
\usepackage[percent]{overpic}

\usepackage{stmaryrd}

\usepackage{amsmath,amsfonts,amssymb}
\usepackage{duckuments}
\usepackage{amsmath}

\usepackage{bm}
\usepackage{nicefrac}
\usepackage{microtype}
\usepackage{dsfont}
\usepackage{changepage}
\usepackage{extramarks}
\usepackage{fancyhdr}
\usepackage{setspace}
\usepackage{soul}
\usepackage{xspace}
\usepackage{hhline}
\usepackage{algorithmicx}
\usepackage{algorithm}
\usepackage{algpseudocode}
\usepackage{pifont}
\usepackage{amssymb}
\usepackage{booktabs}
\usepackage{multirow}

\definecolor{MyDarkBlue}{rgb}{0,0.08,1}
\definecolor{MyDarkGreen}{rgb}{0.02,0.6,0.02}
\definecolor{MyDarkRed}{rgb}{0.8,0.02,0.02}
\definecolor{MyDarkOrange}{rgb}{0.40,0.2,0.02}
\definecolor{MyPurple}{RGB}{111,0,255}
\definecolor{MyRed}{rgb}{1.0,0.0,0.0}
\definecolor{MyGold}{rgb}{0.75,0.6,0.12}
\definecolor{MyDarkgray}{rgb}{0.66, 0.66, 0.66}
\definecolor{MyDarkCyan}{rgb}{0.05, 0.55, 0.45}
\definecolor{MyBlack}{rgb}{0., 0., 0.}
\definecolor{MyMagenta}{rgb}{1., 0., 1.}
\definecolor{BerkeleyYellow}{RGB}{255,204,41}
\definecolor{BerkeleyLightBlue}{RGB}{94,146,221}
\definecolor{BkDarkBlue}{rgb}{.05,.07,.353}

\definecolor{MyDarkGray2}{rgb}{0.6, 0.6, 0.6}

\newcommand{\arxiv}[1]{{#1}}
\newcommand{\camready}[1]{{#1}}

\newcommand*{\menlo}{\fontfamily{lmtt}\fontsize{10}{10}\selectfont }
\newcommand*{\menlofoot}{\fontfamily{lmtt}\fontsize{8}{10}\selectfont }

\newcommand{\suppmat}[1]{#1}
\newcommand{\im}{{\bf x}}

\newcommand{\synthpt}{\widetilde{\im}}

\newcommand{\trainfti}{{\bf t}_{i}}
\newcommand{\synthfti}{{\bf s}_{i}}
\newcommand{\trainftj}{{\bf t}_{j}}
\newcommand{\synthftj}{{\bf s}_{j}}

\newcommand{\ignorethis}[1]{}

\newcommand{\myparagraph}[1]{\vspace{2pt} \noindent \textbf{#1} \ }

\def\1{\bm{1}}

\newcommand{\ignore}[1]{}

\makeatletter
\DeclareRobustCommand\onedot{\futurelet\@let@token\@onedot}
\def\@onedot{\ifx\@let@token.\else.\null\fi\xspace}

\def\etal{\emph{et al}\onedot}
\makeatother

\iccvfinalcopy %

\def\iccvPaperID{1467} %

\ificcvfinal\fi

\begin{document}

\title{Evaluating Data Attribution for Text-to-Image Models}

\author{Sheng-Yu Wang$^{1}$\hspace{5mm} Alexei A. Efros$^{2}$\hspace{5mm} Jun-Yan Zhu$^{1}$\hspace{5mm} Richard Zhang$^{3}$ \\
$^{1}$Carnegie Mellon University \hspace{5mm} $^{2}$UC Berkeley \hspace{5mm} $^{3}$Adobe Research}

\twocolumn[{%
\renewcommand\twocolumn[1][]{#1}%
\maketitle

\begin{center}
    \centering
  
    \includegraphics[width=0.95\linewidth]{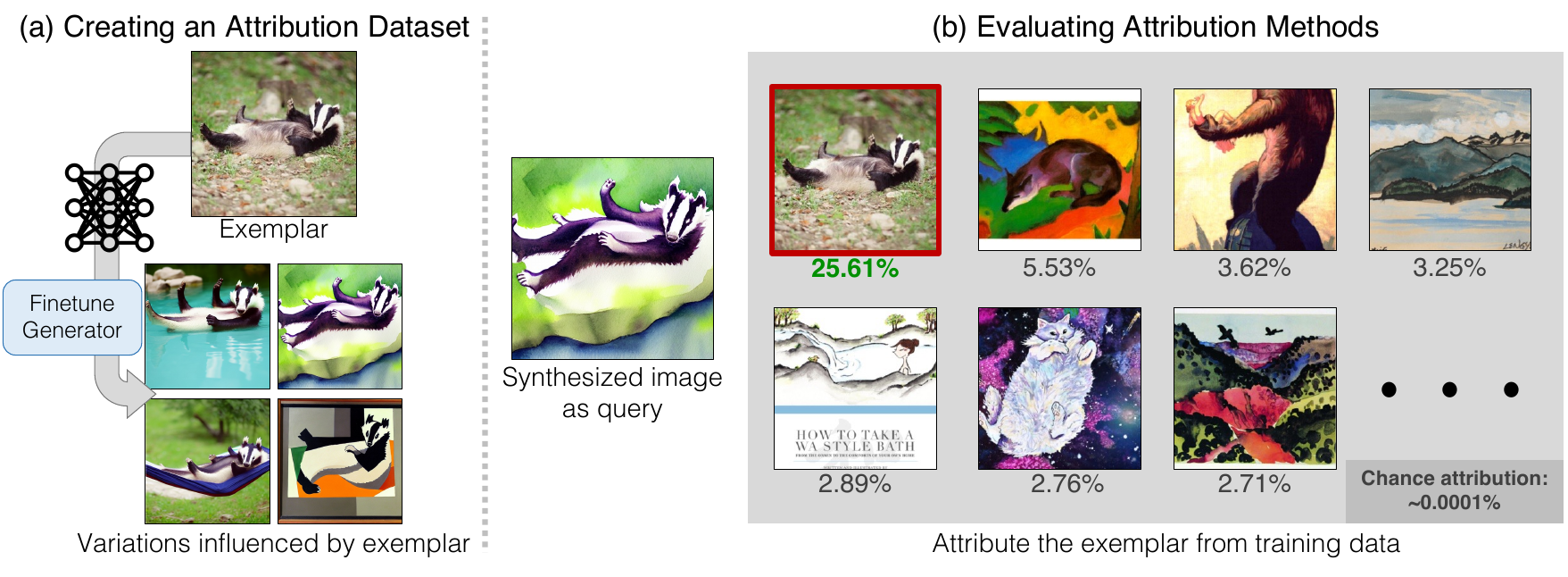}
    \vspace{-13pt}
    \captionof{figure}{
    \textbf{Attribution by Customization (AbC).} (a) We create an visual data attribution     dataset by taking a pretrained generative model and tuning it toward an exemplar image (or images) using ``customization''~\cite{gal2022image,ruiz2022dreambooth,kumari2022customdiffusion}. This produces a set of     synthesized images that are computationally influenced by the exemplar \textit{by construction}.
    (b) Given the dataset, we can evaluate data     attribution approaches by how high they rank the exemplar relative to other training images. Furthermore, using our dataset, we tune representations toward attribution and estimate the probability a given image was an exemplar.}
    \label{fig:teaser}
\end{center}
}]

\maketitle
\ificcvfinal\thispagestyle{empty}\fi

\begin{abstract}
  While large text-to-image models are able to synthesize ``novel'' images, these images are necessarily a reflection of the training data.  The problem of data attribution in such models -- which of the images in the training set are most responsible for the appearance of a given generated image -- is a difficult yet important one.
  \arxiv{As an initial step toward this  problem, we evaluate attribution through ``customization'' methods,
  which tune an existing large-scale model toward a given exemplar object or style.} Our key insight is that this allow us to efficiently create synthetic images that are computationally influenced by the exemplar by construction.
  With our   new dataset of such exemplar-influenced images, we are able to evaluate various data attribution algorithms and different possible feature spaces. Furthermore, by training on our dataset, we can tune standard models, such as DINO, CLIP, and ViT, toward the attribution problem. Even though the procedure is tuned towards small exemplar sets, \arxiv{we show generalization to larger sets}. Finally, by taking into account the inherent uncertainty of the problem, we can assign soft attribution scores over a set of training images.

\end{abstract}

\section{Introduction}

Contemporary generative models create high-quality synthetic images that are ``novel'', i.e., different from any image seen in the training set. 
Yet, these synthetic images would not be possible without the vast training sets employed by the models. 
It is quite remarkable, yet poorly understood, how the generative process is able to compose objects, styles, and attributes from different training images into a coherent novel scene.   Because of copyright and ownership of the training images, understanding the interplay between training data and generative model outputs has become increasingly necessary, both for scientific progress, as well as for practical or legal reasons.

There is a general agreement in the community that not all the billions of training images contribute equally to the appearance of a given synthesized output image.  So, given a particular network output, can we identify the subset of training images that are most responsible for it?
Even for image classifiers, this has remained an open, difficult machine learning problem. Approaches such as influence functions~\cite{koh2017understanding} (inspired by robust statistics), and training and analyzing many models on random subsets~\cite{feldman2020neural} (inspired by Shapley value from economics~\cite{shapley1953value}), are difficult to scale even to modest dataset sizes, let alone the billions of images and parameters that make up modern generative models.

One potential approach that scales well is to run image retrieval in a pre-defined feature space, as, intuitively, synthesized images that are close to a particular training image are more likely to have been influenced by it. However, an out-of-the-box feature space, such as CLIP, is trained for a completely different task and is not necessarily suited for the attribution problem. How can we objectively evaluate which feature spaces are suitable for visual data attribution?

A considerable challenge is obtaining ``ground truth'' attribution data. No method exists for obtaining the set of ground truth training images that influenced a synthesized image.
\arxiv{One way of searching influential images is to check if \textit{removing} particular images during training will affect the model output. This approach for classifiers has been explored by Feldman and Zhang~\cite{feldman2020neural}, where they define influence by training on different subsets of the dataset and analyzing the differences between the resulting models. However, this is computationally infeasible for generation, as training even a single model takes considerable resources, let alone the number of models needed to analyze billions of images. Instead, our work takes inspiration from this but establishes ground truth attribution in a tractable way.}

To do this, we 
exploit a simple insight -- 
by taking a pretrained generative model and tuning it toward a  new exemplar image using ``customization'' methods~\cite{gal2022image,ruiz2022dreambooth,kumari2022customdiffusion}  we can efficiently create synthesized images that are computationally influenced by the exemplar \textit{by construction} (see Figure~\ref{fig:teaser}a). While such synthesized images are not \textit{only} influenced by the exemplar, it serves as a noisy but informative ground truth, and sheds light on how a given training image can be composed into different possible synthesized images.

We create a large dataset of pairs of exemplar and synthesized images using Custom Diffusion~\cite{kumari2022customdiffusion}. We use both exemplar objects (from ImageNet) and styles (from BAM-FG~\cite{ruta2021aladin} and Artchive~\cite{artchive}). Given a synthesized image, the exemplar image, and other random training images from the training set, a strong attribution algorithm should choose the exemplar image over most of the other distractor images.

We leverage this dataset to evaluate candidate retrieval feature spaces, including self-supervised methods~\cite{caron2021emerging,chen2021empirical,radford2021learning}, copy detection~\cite{pizzi2022self}, and style descriptors~\cite{ruta2021aladin}. Furthermore, our dataset can be used to tune the feature spaces to be better suited for the attribution problem through a contrastive learning procedure. Finally, we can estimate the likelihood of a candidate image being the exemplar image by taking a thresholded softmax over the retrieval score. Though our ground truth dataset is of a single image, this enables us to rank and obtain a set of soft attribution scores, assessing ``influence'' over multiple candidate training images. \arxiv{While we train and evaluate for attribution on exemplar-based customization (1-10 related images), we demonstrate that the method generalizes even when tuning on larger sets (100-1000 random, unrelated images), suggesting applicability to the general, more challenging data attribution problem.}

To summarize our contributions: 1) We propose an efficient method for generating a dataset of synthesized images paired with ground-truth exemplar images that influenced them.
2) We leverage this dataset to evaluate candidate image retrieval feature spaces. Furthermore, we demonstrate our dataset can improve feature spaces through a contrastive learning objective.
3) We can softly assess influence scores over the training image dataset. Our code, model, and dataset are released at: \url{https://peterwang512.github.io/GenDataAttribution}.

\begin{figure*}
    \centering
    \includegraphics[width=0.95\linewidth]{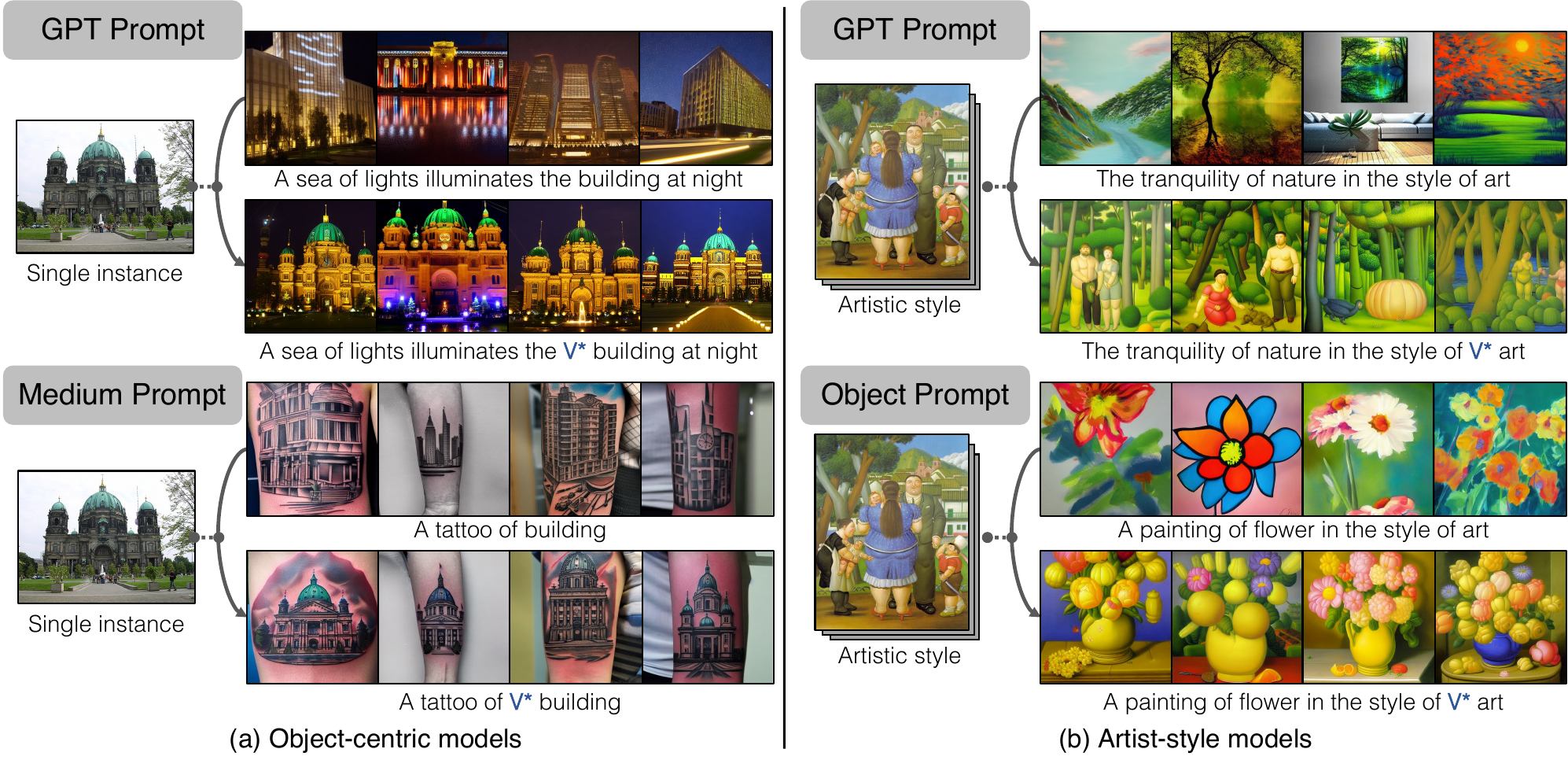}
    \vspace{-12pt}
    \caption{\textbf{Example synthesized images.} We show qualitative examples of our Custom Diffusion datasets. We generate object-centric models (left) and artistic style models (right). For each, we use ChatGPT (top) to generate prompts and procedurally generate prompts (bottom).}
    \label{fig:prompting}
    \vspace{-15pt}
    
\end{figure*}

\section{Related Work}

\myparagraph{Deep generative models.}
Deep generative models aim to learn a distribution from a training set~\cite{goodfellow2014generative,kingma2014auto,dinh2017density,oord2016conditional,razavi2019generating,esser2021taming}. %
Recently, diffusion models~\cite{sohl2015deep,song2020denoising,ho2020denoising,song2020score,dhariwal2021diffusion}
have become the de facto method for generating images. Adapting such models to text-to-image synthesis~\cite{rombach2021highresolution,nichol2021glide,nichol2021improved} has produced a rash of models of amazing quality and diversity, including Imagen~\cite{saharia2022photorealistic} and DALL$\cdot$E-2~\cite{ramesh2022hierarchical}. Such models can serve as a high-quality image prior for image editing~\cite{brooks2022instructpix2pix,hertz2022prompt,kawar2022imagic,tumanyan2022plug,parmar2023zero,meng2022sdedit}. Additionally, of special interest to our attribution method is algorithms that quickly customize a pretrained model towards an additional exemplar or concept~\cite{ruiz2022dreambooth,gal2022image,kumari2022customdiffusion}. Though our method applies to all model types at a high level, we specifically study the open-source Stable Diffusion model~\cite{rombach2021highresolution}, as we have direct access to the model.

\myparagraph{Model attribution.} %
Given a synthetic image, recent works seek to identify which model they came from~\cite{yu2019attributing,bui2022repmix,sha2022fake,marra2019gans} or can represent it~\cite{lu2022content}, known as \text{model} attribution. From there, identifying what images are in the training set of the model is known as membership inference (attack). This is an open problem, actively being explored for both discriminative~\cite{sablayrolles2018d,shokri2017membership,hu2022membership} and generative models~\cite{hayes2017logan,hilprecht2019monte,chen2020gan,carlini2021extracting}.
In our work, we assume the origin of a synthetic image is known and take this a step further. We aim to attribute the \textit{specific} training data elements that made it possible.

\myparagraph{Influence estimation in discriminative models.} An important related line of work is the field of influence functions, seeking to explain how each training point affects a model's output. In their seminal work, Koh and Liang~\cite{koh2017understanding} explore this in the context of discriminative models, estimating the influence of up- or down-weighting a point on the objective function. This requires creating a Hessian the size of the parameters, and later work seeks to make the problem more tractable~\cite{khanna2019interpreting,guo2020fastif,schioppa2022scaling,akyurektowards,park2023trak}. 

Several works aim to measure the Shapley value of datapoints~\cite{shapley1953value}, a landmark concept from economics and game theory, which is the average expected marginal contribution of one player after all possible combinations have been considered
~\cite{ghorbani2019data,jia2019towards}. For example, Feldman and Zhang~\cite{feldman2020neural} train on random subsets, assessing a training point by computing the objective score of models with and without the point. Pruthi et al.~\cite{pruthi2020estimating} explore tracing the loss function as the model sees each training sample, positing that decreases in training objective on a test point explains the causal effect of a given training point. %
These works focus on discriminative models, whereas we work with generative ones.

Unfortunately, training large generative models even once is prohibitively expensive (except for the most well-funded organizations), let alone the many times required to estimate the value of each training point. Our method is inspired by these previous works, aiming to overcome the fundamental difficulties in translating them into the space of large generative models. \arxiv{Instead of analyzing training influences post-hoc, which is currently intractable, our key idea is to simulate ground truth directly by adding exemplar influences.}

\myparagraph{Replication detection.} A special case where training images have an outsized influence is if a synthesized image is a ``copy''. Even in human-created art, common patterns can be reused, for example, common elements in paintings~\cite{shen2019discovering}, or animation patterns in movies~\cite{disney}. As exact bit-wise matches are unlikely to occur and the space of synthesized and training images is large, several works~\cite{somepalli2022diffusion,carlini2023extracting} aim to efficiently mine for approximate matches. In such special cases, influence is near $100\%$ for a single training image. Our work aims to assess the influence over the whole training set in general cases even when the synthesized image is not a direct copy.

\myparagraph{Representation learning.}
Advances in self-supervised learning have produced strong representations, learned from large-scale data with weak or no supervision~\cite{radford2021learning,caron2021emerging,he2020momentum,tian2020contrastive,chen2020simple,misra2019self}. 
These advancements have produced feature representations that are useful for downstream recognition tasks~\cite{he2017mask} 
We study if such candidate image representations are suited for the problem of attribution and improve them by using contrastive learning on our data.

\begin{table*}[h!]
    \centering
    \resizebox{.9\linewidth}{!}{
    \begin{tabular}{llcccccccccc}
    \toprule
    & & \multicolumn{5}{c}{Object-centric} & \multicolumn{5}{c}{Artistic styles} \\ \cmidrule(lr){3-7} \cmidrule(lr){8-12}
    \multicolumn{2}{c}{\multirow{2}{*}{Property}} & \multicolumn{3}{c}{Imagenet-Seen} & Unseen & \multirow{2}{*}{Total} & \multicolumn{3}{c}{BAM-FG} & Artchive & \multirow{2}{*}{Total}  \\ \cmidrule(lr){3-5} \cmidrule(lr){6-6} \cmidrule(lr){8-10} \cmidrule(lr){11-11}
    
    & & train  & val & test & test & & train     & val   & test   & test & \\ \midrule %
    
    \multicolumn{2}{c}{Object classes} & 593   & 593   & 593         & 100$^{*}$ & 693 & -- & -- & -- & -- & -- \\
    \multicolumn{2}{c}{Training images} & 4744   & 593   & 593         & 1000 & 6930 & 78,086 & 1837  & 1692   & 3081 & 84,696 \\ \cdashline{1-12}
    
    \multicolumn{2}{c}{Avg images/model} & 1 & 1 & 1 & 1 & 1 & 7.36 & 7.35 & 6.77 & 12.08 & 7.45 \\
    \multicolumn{2}{c}{Total models} & 4744   & 593   & 593         & 1000 & 6930 & 10,607 & 250   & 250    & 255 & 11,362 \\ \cdashline{1-12}
    
    \multirow{2}{*}{Prompts} & ChatGPT$^\dagger$ & 15 & 6 & 10 & 10 & -- & 40 & 6 & 10 & 10 & 50 \\  
    & Procedural & 40 &  6 & 10$^\ddag$ & 10$^\ddag$ & 50 & 30 & 6 & 10$^\ddag$ & 10$^\ddag$ & 40 \\ \cdashline{1-12}
    \multirow{3}{*}{Samples} & ChatGPT & 284,640 & 14,232 & 23,720 & 40,000 & 362,592 & 1,697,120   & 6,000  & 10,000  & 10,200 & 1,723,320 \\
    & Procedural & 759,040 & 14,232 & 23,720 & 40,000 & 836,992 & 1,272,840 & 6,000 & 10,000 & 10,200 & 1,299,040 \\ \cdashline{2-12}
    & Total & 1,043,680 & 28,464 & 47,440 & 80,000 & 1,199,584 & 2,969,960 & 12,000 & 20,000 & 20,400 & 3,022,360 \\
    \bottomrule
    \end{tabular}
    }
    \vspace{-.1in}
    \caption{\textbf{Dataset statistics.} We tune Custom Diffusion models on objects and artistic sets. In total, we train $>$18,000 models and draw $>$4M samples. To draw samples, we use ChatGPT for each category and procedural generation (mixing object-centric models with a bank of different styles, and style-tuned models with a bank of objects) to generate prompts. Note that we take special care to create distinct training and testing distributions. $^{*}$ We avoid overlapping ImageNet classes for objects and overlapping data sources for artistic models. We use a distinct set of prompts for testing. $^\dagger$ChatGPT-generated prompts are per-broad category (e.g., cat, dog, etc.) for objects and shared across artistic styles. $^\ddag$Distinct, procedural prompts are used for out-of-distribution models.
    }
    \label{tab:dataset}
    \vspace{-15pt}
    
\end{table*}

\section{Creating an attribution dataset}
\label{sec:dataset}

We take the first step to define, evaluate, and learn data attribution for large-scale generative models.
Given a dataset $\mathcal{D} = \{({\bf x}, {\bf c})\}$, containing images ${\bf x} \in \mathcal{X}$ and conditioning text $ {\bf c}$, the generative model training process seeks to train a generator $G: ({\bf z}, {\bf c}) \rightarrow {\bf x}$, where ${\bf z} \sim \mathcal{N}(0, {\bf I})$ and sampled image ${\bf x}$ is drawn from the distribution $p({\bf x} | {\bf c})$. We denote $\mathcal{X}$ as the original training set (e.g., LAION) and the training process as $T: \mathcal{D}\rightarrow G$.
\subsection{``Add-one-in'' training}
\label{sec:formulation}

We perturb the training process by training a generator with an added concept, which can contain one or several images and one prompt $\mathcal{D}^{+} = \{ ({\bf x}^{+}, {\bf c}^{+}) \}$. \arxiv{This produces a new generator: $G^{+} = T^{+} \big(\mathcal{D}, \mathcal{D}^{+} \big)$, where $T^{+}$ stands for pre-training on $\mathcal{D}$ and then fine-tuning on $\mathcal{D}^{+}$. 
}

As training a new generator from scratch for each concept would be prohibitively costly, we use Custom Diffusion~\cite{kumari2022customdiffusion}, an efficient fine-tuning method (6 minutes) with low storage requirements (75MB). This method presents an efficient approximation in terms of runtime, memory, and storage and enables us to scale up the collection of models and images in a tractable manner. We sample from the new generator:
\begin{equation}
    \widetilde{{\bf x}} = G^{+}({\bf z}, \mathcal{C}({\bf c}^{+})), %
\end{equation}

\noindent where function $\mathcal{C}$ represents the prompt engineering process for generating a random text prompt related to the concept ${\bf c}^{+}$. 
We also denote the exemplar set as $\mathcal{X}^{+}= \{ {\bf x}^{+} \}$ and the synthesized images set as $\widetilde{\bf \mathcal{X}} = \{ \widetilde{{\bf x}} \}$. %
We describe the dataset selection process next. %

\myparagraph{Dataset curation.} With infinite compute, we could exhaustively sample from the original training set $\mathcal{D}$ and sample infinite, random prompts from those models. 
However, this approach is not computationally tractable for large datasets such as LAION 5B~\cite{schuhmann2022laion}.  
Also, random images in LAION can contain arbitrary, noisy prompts\footnote{For example, ``\menlofoot{You Won't Believe How These 10 Famous Companies Started Out [INFOGRAPHIC]}''}, 
making identifying related prompts non-trivial. \arxiv{Therefore, to create a clean dataset, we choose a set of images for which we can identify clean concept names and designate related prompts.}

We create two groups of models to measure different aspects of generation: (1) Object-centric models: we add \arxiv{an exemplar} object instance of a known class, using a single training exemplar. (2) Artistic style-centric models: we add \arxiv{an exemplar} style defined by a small image collection. For each group, we create two separate sets, enabling us to test \textit{out-of-distribution generalization} when tuning representations on our dataset.

Figure~\ref{fig:prompting} shows samples of our dataset and prompts, and Table~\ref{tab:dataset} summarizes the dataset statistics. Next, we describe our dataset composition and prompt engineering strategies.

\subsection{Object-centric models}
\label{sec:object_models}
We use the validation set of ImageNet-1K~\cite{deng2009imagenet}, a clean choice for training customized models and generating prompts, thanks to its annotated class labels and highly diverse categories. 
During training, we take a single image of a given category %
and train with the text prompt ``{\menlo V$^{*}$ cat}'', where {\menlo cat} is the broader category of the image and {\menlo V$^{*}$} is a token, used by Custom Diffusion~\cite{kumari2022customdiffusion} to associate to the input exemplar. We train each Custom Diffusion model on a \textit{single} exemplar image and only use the category names for prompt engineering during training and synthesis time. \arxiv{We also manually remove categories where Custom Diffusion does not generate samples that match the exemplar faithfully.}

As summarized in the first row of Table~\ref{tab:dataset}, we select 6930 images from 693 ImageNet classes (10 images/class), where customized models generate the inserted concept more faithfully. Of these, we build two sets -- (1) Seen classes: we select 5930 images from 593 classes,  with the images divided to train (80\%), val (10\%), and test (10\%).  %
The train and val set can be later used to tune attribution models. (2) Unseen classes: to facilitate out-of-distribution testing, we hold out the classes from ImageNet-100, each containing 10 instance models, making 1000 models total.

\myparagraph{Prompt-engineering for objects.} Next, we leverage the fine-tuned models $G^+$ to generate images related to the inserted concept. We use two methods to generate the inserted concept, with a diverse set of scenarios. First, we query ChatGPT~\cite{chatgpt} for prompts containing the object instance. Such prompts will generate the object in different poses, locations, or performing different actions, e.g., ``{\menlo The V$^{*}$ cat groomed itself meticulously}''. 

The ChatGPT-based method generates diverse prompts that would be difficult to hand-query for hundreds of classes. However, such prompts tend to retain the same photorealistic appearance, while text-to-image models can also generate concepts in different stylized appearances. As such, we procedurally generate prompts for the object depicted in different mediums. For example, ``{\menlo A <medium> of V$^{*}$ cat}'', using mediums such as ``{\menlo watercolor painting, tattoo, digital art}'', etc.

For each model, we have 12-60 prompts, with 3-4 samples/prompt, resulting in 80-220 synthesized images per model. Details are shown in Table~\ref{tab:dataset}. In total, we generate $>1$M training images and $47,440$ and $80,000$ for in and out-of-distribution testing, respectively. Note that we use separate prompts for the out-of-distribution test set. %

\subsection{Artistic-style models}
To train artistic-style models, we use images from two sources: (1) BAM-FG (Behance Artistic Media - Fine-Grained dataset)~\cite{ruta2021aladin} -- a dataset drawn from groups of images from Behance, validated to be of the same ``style'' grouped by users. We select the subset of BAM-FG with the highest user consensus. %
(2) Artchive~\cite{artchive}, a website with collections of paintings by well-known artists, such as Cezanne, Botero, and Millet. \arxiv{We train each model on images of the same style or by the same artist, with the text prompt ``A picture in the style of {\menlo V$^{*}$} art''.}
In total, we collect 11,107 models from BAM-FG and 255 models from Artchive, with each model tuned on 7.35 and 12.1 images on average, respectively. We split the BAM-FG models into 10,607 for training, 250 for validation, and 250 for testing. All 255 Artchive models are used for testing.

\myparagraph{Prompt-engineering for styles.} We aim to generate synthetic images with high diversity, in relation to the concept style.
First, we use ChatGPT to sample 50 painting captions and create prompts such as ``{\menlo The magic of the forest in the style of V$^{*}$ art}''. To generate diverse objects, we specify 40 different objects, such as flowers and rivers, to procedurally prompts in the form of ``{\menlo A picture of <object> in the style of V$^{*}$ art}'' for BAM-FG or ``{\menlo A painting...}'' for Artchive models.

For testing, we hold out 10 prompts from each prompting scheme and use the rest for training and validation. As Artchive is from a different data source, and these prompts are distinctly held out, this serves as an out-of-distribution test set when training on the BAM-FG data.
In practice, we use a subset of training prompts for validation.

\myparagraph{Summary.}
In total, we generate $>$1M images from 7000 object-centric models and almost 3M images from $>$11,000 artistic style models. We split these into out-of-distribution test sets, by using different data sources and held-out prompts. \arxiv{We also manually select plausible ChatGPT-generated prompts during data curation. \suppmat{We provide example ChatGPT queries, a list of our prompts, and more detailed dataset statistics in Appendix~\ref{sec:supp_data}.}} Next, we evaluate different feature spaces and then improve the features for attribution. %

\section{Evaluating and training for attribution}
\label{sec:learning}

We have defined a process for creating a dataset of $N$ models, containing paired sets of exemplar training images $\mathcal{X}^{+}_i$ and influenced synthetic images $\widetilde{\mathcal{X}}_i$, %
taking into account the generative modeling training process.
Next, we aim to learn the inverse -- predicting corresponding influencing set $\mathcal{X}^{+}_i$ from a generated image $\widetilde{\bf x} \in \widetilde{\mathcal{X}}_i$. %

\subsection{Evaluating existing features}
\vspace{-.05in}
\label{sec:eval_nn}

We first note that a feature extractor $F$ well-suited for attribution would place images in $\mathcal{X}^{+}_i$ with higher similarity to $\widetilde{{\bf x}}$, as compared to the other random images in $\mathcal{X}$, which is the dataset used to pretrain the model:%

\begin{figure}
    \centering
    \includegraphics[width=\linewidth]{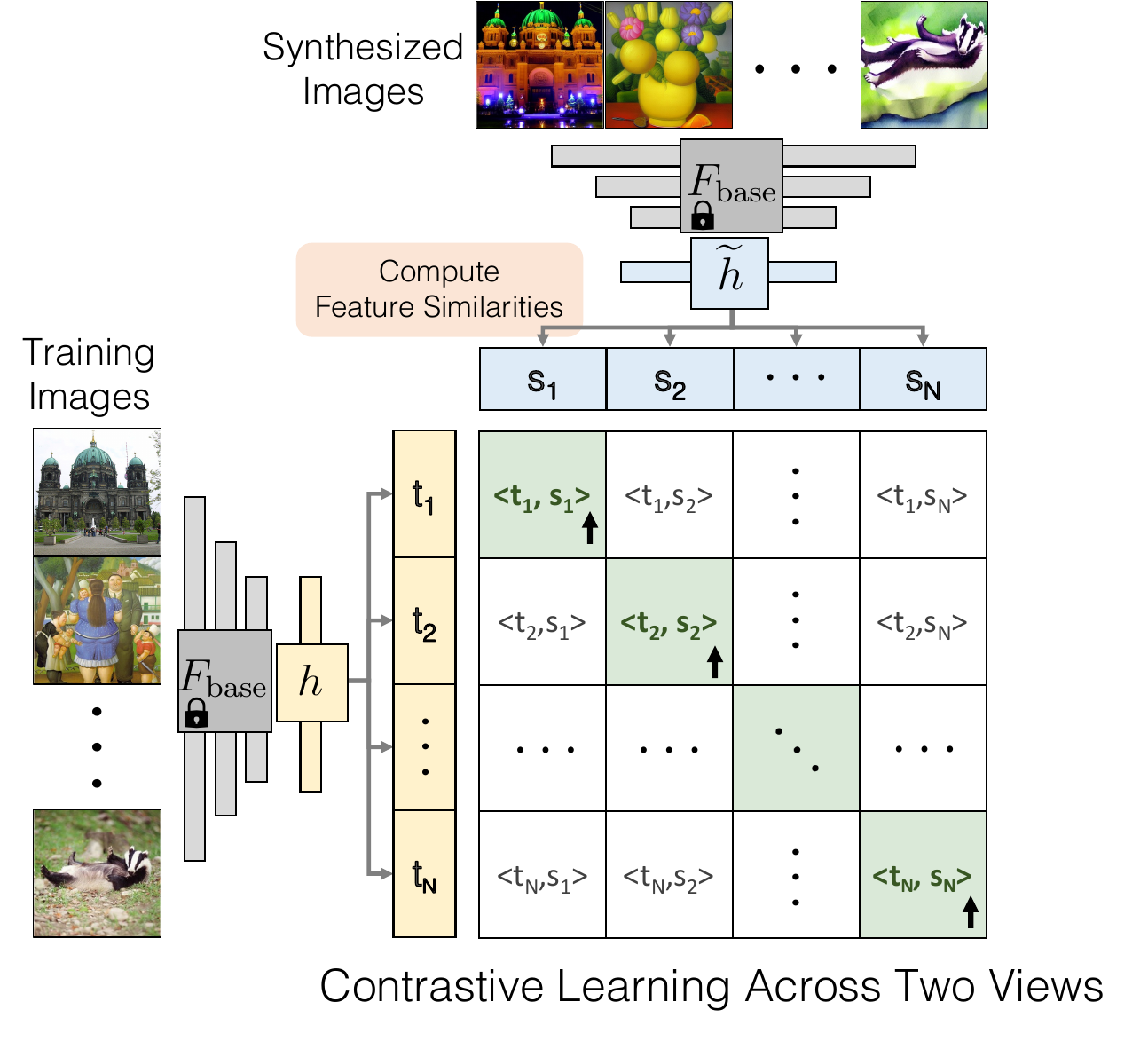}
    \vspace{-25pt}
    \caption{\textbf{Training for attribution.} We use contrastive learning to learn a linear layer $h, \widetilde{h}$ on top of an existing feature space $F_\text{base}$, using our attribution dataset. The embedding learns high similarity for corresponding training and synthesized images, in contrast to non-corresponding images from the dataset. }
    \label{fig:method}
    \vspace{-15pt}
\end{figure}

\begin{equation}
    \text{sim} \big( F({\bf x}^+), F(\widetilde{\bf x}) \big) > \text{sim} \big( F({\bf x}^{-}), F(\widetilde{\bf x}) \big),
    \label{eqn:sim}
\end{equation}
\noindent where $\widetilde{\bf x} \in \widetilde{\mathcal{X}}_i$, ${\bf x}^+ \in \mathcal{X}^{+}_i$ and ${\bf x}^{-} \in  \mathcal{X} \hspace{-.5mm}  $. Again, $\mathcal{X}$ denotes the original training set (e.g., LAION). 
This allows us to evaluate data attribution with standard information retrieval protocols. In Section~\ref{sec:exp}, we evaluate candidate feature spaces (e.g., CLIP~\cite{radford2021learning}, DINO~\cite{caron2021emerging}), to see which ones are more suited out-of-the-box for data attribution. While these features serve as a reasonable baseline and perform well above chance, assessing visual similarity is not equivalent to attributing data influence. Next, we learn a better function for data attribution by finetuning pretrained features on our dataset.

\subsection{Learning features for attribution}
\vspace{-.05in}
\label{sec:attribute_nn}
Our dataset consists of paired views of training and synthesized sets. This lends itself naturally to contrastive learning~\cite{oord2018representation,tian2020contrastive} to capture the association between the two views.

\myparagraph{Contrastive learning.}
We apply a frozen, pretrained image encoder $F_\text{base}$ along with light mapping functions $h$, $\tilde{h}$, creating feature extractors $F=h \circ F_\text{base}$, $\widetilde{F}=\widetilde{h} \circ F_\text{base}$.
We apply the commonly used NT-Xent (normalized temperature cross-entropy) loss~\cite{chen2020simple}:
\begin{equation}
    \mathcal{L}_\text{cont}^i = - \Big( \log \frac{\exp( \trainfti^\top \synthfti / \upsilon)}{\sum_{j} \exp( \trainfti^\top \synthftj / \upsilon)}
    + \log \frac{\exp( \trainfti^\top \synthfti / \upsilon)}{\sum_{j} \exp( \trainftj^\top \synthfti / \upsilon)} \Big),
    \label{eqn:contrastive}
\end{equation}

\noindent where $\trainfti=F({\bf x}^{+})$ and $\synthfti=\widetilde{F}({\widetilde{\bf x}})$ are normalized features extracted from training and synthesized images, respectively, and $\trainftj$, $\synthftj$ are extracted features from images in the dataset.
\arxiv{The loss encourages feature similarity $\trainfti^\top \synthftj$ to be large for positive pairs $(i = j)$ and small otherwise $(i \neq j)$.}
Here, negatives are drawn from other exemplar images $\cup_{j\neq i} \mathcal{X}^{+}_j$, rather than the original LAION dataset. With LAION negatives, the network will learn to classify datasets (e.g., ImageNet vs. LAION), rather than learning attribution.  %
We set the temperature $\upsilon = 1$ during training.

\myparagraph{Regularization.} We experiment with different mapping functions and find that affine mapping performs the best. We denote the affine mappings as $H({\bf x}) = W{\bf x} + b$ and $\widetilde{H}({\bf x}) = \widetilde{W}{\bf x} + \widetilde{b}$, where $W, \widetilde{W}$ are square matrices. We find that directly optimizing with $\mathcal{L}_\text{cont}$ leads to overfitting, so we regularize the mapping and add it to define the following attribution loss:
\begin{equation}
    \begin{aligned}
        \mathcal{L}&_\text{attribution} = \mathds{E}_i [\mathcal{L}_\text{cont}^i] +  \lambda_\text{reg} \mathcal{L}_\text{reg}, \\
        \text{where } &\mathcal{L}_\text{reg} = \frac12\left(||W^\top W - I||_F + ||\widetilde{W}^\top \widetilde{W} - I||_F \right).
    \end{aligned}
\end{equation}

\noindent We set $\lambda_\text{reg} = 0.05$. Figure~\ref{fig:method} summarizes the learning procedure. \suppmat{More training details are in Appendix~\ref{sec:supp_mapper_train_detail}.}

\myparagraph{Extracting soft influence scores.}
\arxiv{From the learned feature similarities, we obtain a soft probabilistic influence score $\hat{P}({\bf x} | \widetilde{\bf x}; \mathcal{X}^{+} \cup \mathcal{X})$, which indicates how likely a candidate ${\bf x}$ is in the influencing set $\mathcal{X}^{+}$ for synthetic content $\widetilde{\bf x}$.
The ground truth influence is split amongst the exemplars 
$\mathcal{S}({\bf x}; \synthpt) = \tfrac{1}{|\mathcal{X}^{+}|} \mathds{1}_{\{ {\bf x} \in \mathcal{X}^+ \}}$.
We optimize the KL divergence:
}
\vspace{-5pt}
\begin{equation}
    \begin{aligned}
        \min_P \hspace{1mm} \mathbb{E}_{ \hspace{.5mm}{\synthpt}} \hspace{1mm} \mathcal{D}_\text{KL} \hspace{-.5mm} \left[ \mathcal{S}({\bf x}; \synthpt) \hspace{.5mm} || \hspace{.5mm} \hat{P}({\bf x} |\synthpt; \mathcal{X}^{+}  \cup \mathcal{X}) \right]
    \end{aligned}
    \label{eqn:kl_calib}
\end{equation}

\arxiv{
This resembles our retrieval objective (Equation~\ref{eqn:contrastive}). As such, images retrieved with higher similarities $s \equiv F({\bf x})^{\top} \widetilde{F}(\widetilde{\bf x})$ should have higher score. Our job is simply to monotonically map similarity scores, ordered as $s_{(0)}\geq s_{(1)}\geq \dots\geq s_{(|\mathcal{X}^+ \cup \mathcal{X}|)}$, to well-calibrated probabilities. To do this, we use an exponential (with learned temperature $\tau$), ReLU (with learned offset $\lambda$), and normalize.
}

\begin{equation}
    \begin{aligned}
        \hat{P}_{\tau,\lambda}({\bf x} |\synthpt; \mathcal{X}^{+}  \cup \mathcal{X}) =  \frac{ \text{ReLU} \big( \exp[ \sfrac{(s - s_{(0)})}{\tau}] - \lambda \big)}{ \sum_j \text{ReLU}\big( \exp [ \sfrac{(s_{(j)} - s_{(0)})}{\tau} ] - \lambda \big) },
    \end{aligned}
    \label{eqn:calib_score_func}
\end{equation}

\begin{figure}[t!]
    \centering
    \includegraphics[width=\linewidth]{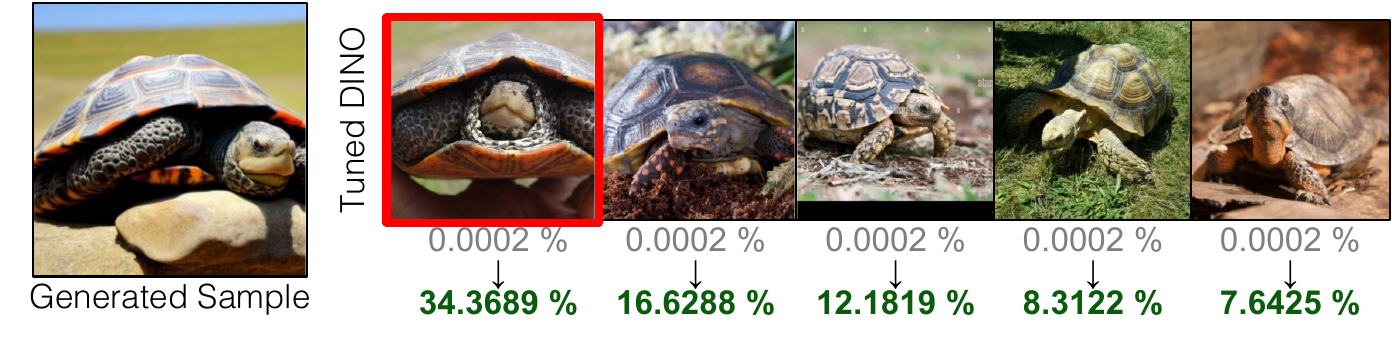}
    \vspace{-20pt}
    \caption{{\bf Calibration effects.}
    {\color{gray} \textbf{Gray}} is the original softmax scores without calibration, and {\color{MyDarkGreen} \textbf{green}} is after our calibration. Here influence scores based on DINO features tuned with our full dataset.}
    \label{fig:calib_effect}
    \vspace{-15pt}
    
\end{figure}

\arxiv{Note that without the ReLU, this is simply a softmax function. The ReLU enables the optimization to learn to assign zero probability to low-influence images. During optimization, we approxmiate ReLU with softplus for training stability.  We also subtract the similarity score by the maximum $s_{(0)}$ to apply thresholding at a similar scale across instances. Figure~\ref{fig:calib_effect} demonstrates the effectiveness of our calibration, relative to default temperature $\tau=1$ and no thresholding. \suppmat{Additional details are in Appendix~\ref{sec:supp_soft}.}}

\begin{figure*}
    \centering
    \begin{tabular}{cc}
        \includegraphics[width=0.42\linewidth]{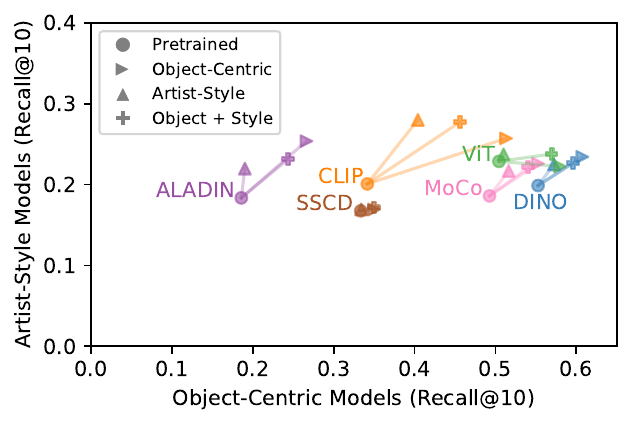} & 
        \includegraphics[width=0.5\linewidth]{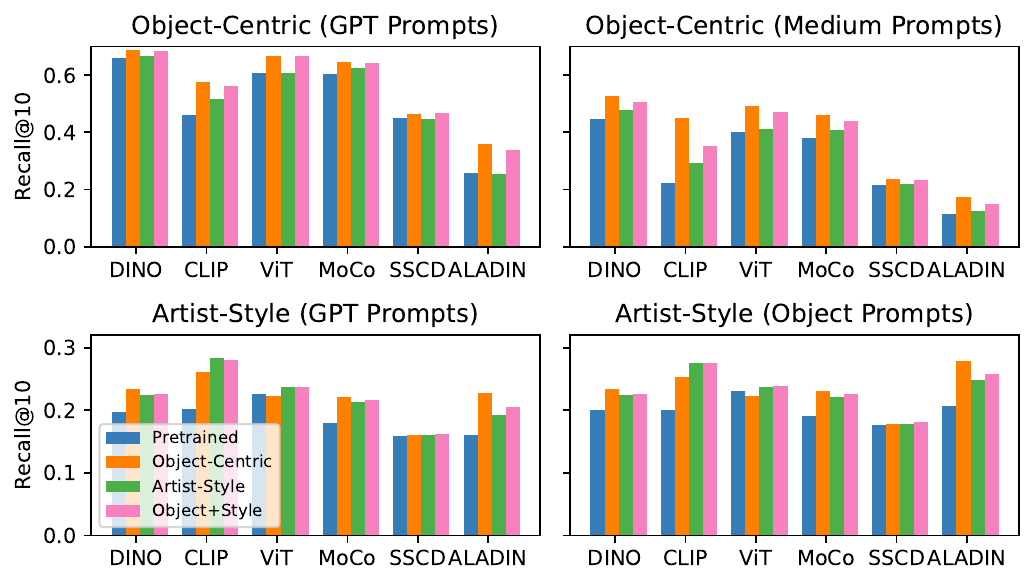} \\
    \end{tabular}
    \vspace{-15pt}
    \caption{\textbf{Quantitative comparison.} We show the performance of pretrained models and trained models on our attribution dataset. (Left) We show performance on artistic-style models on (y-axis) vs. object-centric models (x-axis). We first evaluate pretrained feature spaces. Next, we show how training on object-centric and artistic-style models separately or together, often leads to improvements on each axis. %
    All three variants of CLIP and the object-centric model of DINO lie on the Pareto front. (Right) We show the performance, broken up by object-centric vs. artistic-style datasets and prompting procedures. DINO, CLIP, ViT, and MoCo significantly outperform copy detection SSCD and style-descriptor ALADIN. We observe consistent performance gains across prompt types when training on our dataset.}
    \label{fig:bar_plot}
    \vspace{-15pt}
    
\end{figure*}

\myparagraph{Summary.}
Our data curation process (Section~\ref{sec:dataset}) is a forward influence-generation process, generating synthesized images %
through the customization process $T^{+}$.
In this section, we learn to reverse -- first by training a feature extractor to \textit{retrieve} $\mathcal{X}^{+}$ from $\widetilde{\bf x}$. Then, by taking a calibrated softmax over the similarities scores, we estimate probability \arxiv{$\hat{P}_{\tau,\lambda}$} for which images were in the exemplar training set. %
\begin{figure}[t!]
    \centering
    \includegraphics[width=\linewidth]{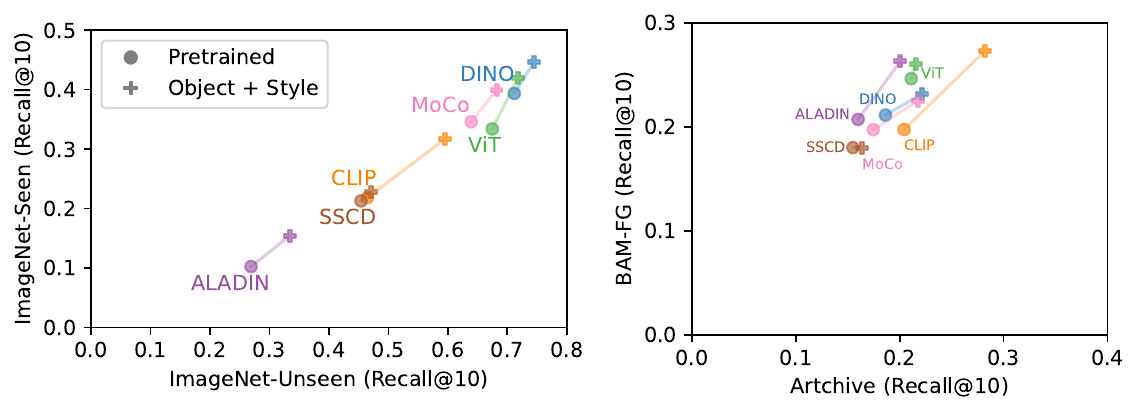}
    \vspace{-20pt}
    \caption{\textbf{In/out-of-domain tests.} Training on one distribution (Imagenet-Seen and BAM-FG), plotted on the x-axes, generalizes to an unseen distribution -- Imagenet-Unseen (left) and Artchive (right) -- plotted on the y-axes. Almost all points move up and to the right, indicating successful out-of-domain generalization.}
    \label{fig:outdomain}
    \vspace{-5pt}
    
\end{figure}

\begin{figure}[t!]
    \centering
    \includegraphics[width=\linewidth]{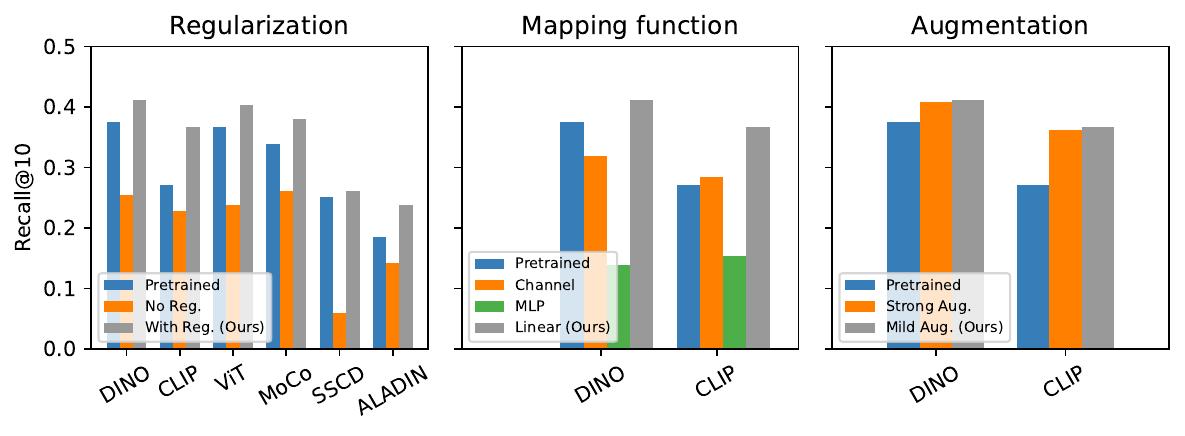}
    \vspace{-25pt}
    \caption{\textbf{Ablation studies.} We study the effects of different design choices, averaged across all test cases. (Left) Training with regularization is critical for improving performance. (Middle) Linear layer mapping works the best. %
    (Right) Mild augmentation performs slightly stronger than strong augmentation.}
    \label{fig:ablation}
    \vspace{-10pt}
    
\end{figure}

\section{Experiments}
\label{sec:exp}

\begin{figure*}
    \centering
    \includegraphics[width=1.\linewidth]{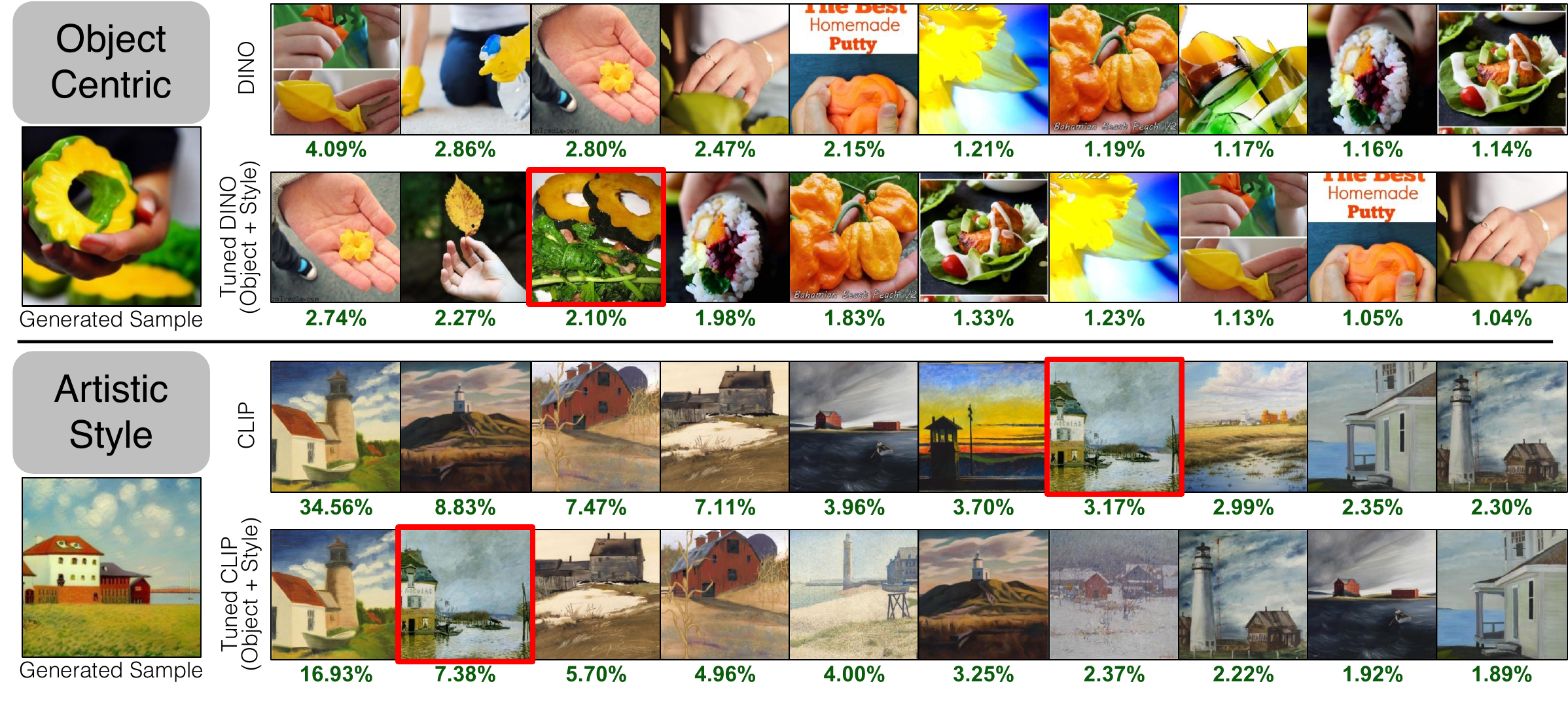}
    \vspace{-25pt}
    
    \caption{\textbf{Qualitative comparison.} For a given synthesized sample, obtained by training on an image of an acorn squash (top) and paintings by Alfred Sisley (bottom), our fine-tuned attribution method improves the ranking and influence score of the exemplar training image. %
    }
    \label{fig:main_qual}
    \vspace{-5pt}
    
\end{figure*}

\definecolor{gggreen}{RGB}{50, 128, 30}
\begin{figure*}
    \centering
    \includegraphics[width=1.\linewidth]{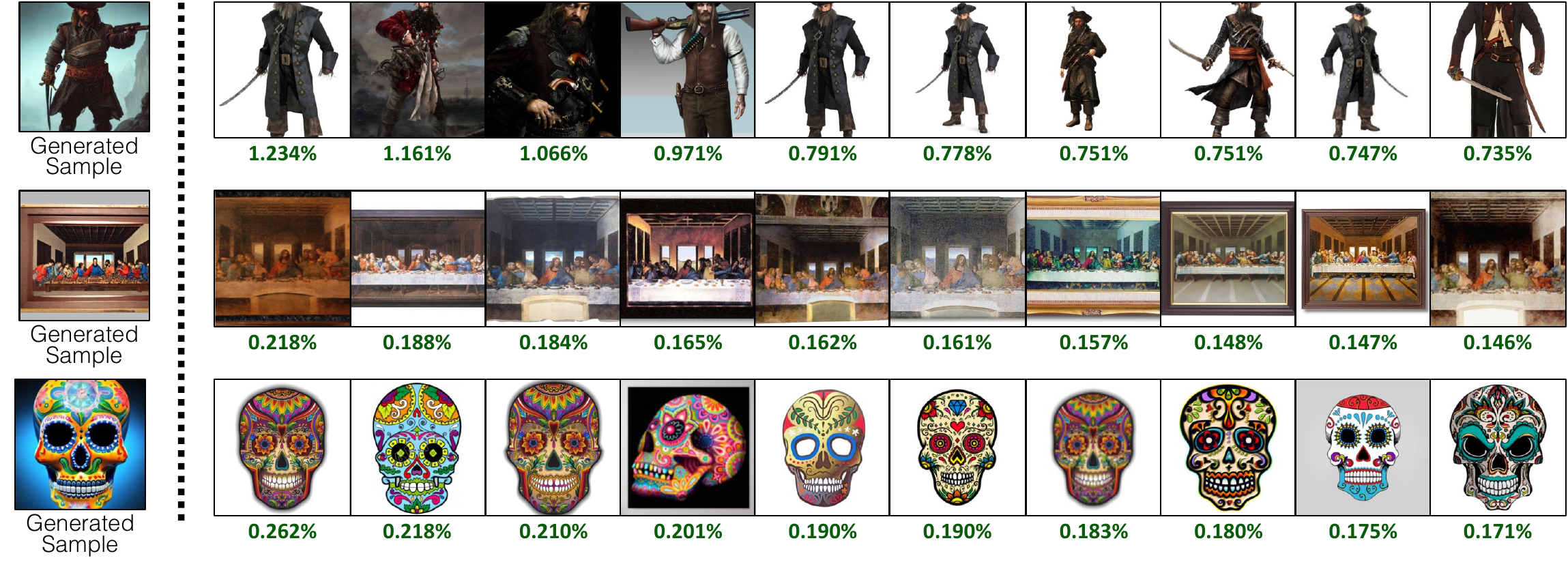}
    \vspace{-25pt}
    
    \caption{\textbf{Attributing Stable Diffusion Images.} We run our influence score prediction function with CLIP, tuned on our Object+Style attribution datasets. In each row, we show a generated sample query (Left), and the top attributed training images from LAION-400M (Right). {\bf {\color{gggreen} Green}} values are calibrated influence percentage scores.}
    \label{fig:sd_nn}
    \vspace{-10pt}
    
\end{figure*}

\myparagraph{Metrics and test cases.}
We evaluate attribution with two metrics: \textbf{(1) Recall@K}: proportion of influencers $\mathcal{X}^{+}$ in top-K retrieved images, \textbf{(2) mAP}: a ranking-based score to evaluate the overall ordering of retrieval. To evaluate our method efficiently, we retrieve from a union of the added concepts and a random 1M subset of LAION-400M. We have 8 test cases \arxiv{separated from our train and val set}, each with 2 prompting schemes per split and 4 different test splits. We calculate the average metrics over queries for each test case and average the metric across test cases when reporting numbers for broader categories such as style-centric models.

We test several image encoders, including self-supervised (DINO~\cite{caron2021emerging}, MoCo v3~\cite{chen2021empirical}), language-pretrained (CLIP~\cite{radford2021learning}), supervised (ViT~\cite{dosovitskiy2021image}), style descriptor (ALADIN~\cite{ruta2021aladin}), and copy detection (SSCD~\cite{pizzi2022self})  methods. For DINO, MoCo, CLIP, and ViT, we use the same ViT-B/16 architecture for a fair comparison. We evaluate the encoded features, with and without our learned linear mapping. We train mappers with (1) object-centric models only, (2) style-centric models only, and (3) both. \suppmat{We report R@10 in the main text and include other metrics in Appendix~\ref{sec:supp_additional_metrics}, as trends are consistent across metrics.} Figure~\ref{fig:bar_plot} shows results for different types of customized models and prompting schemes, and Figure~\ref{fig:main_qual} shows performance in object-centric and style-centric models, separately across different models.

\myparagraph{What feature encoder to start with?}
We study which features are suitable for data attribution. Figure~\ref{fig:bar_plot} shows that ImageNet-pretrained encoders (DINO, ViT, MoCo) perform better in object-centric models. This indicates a smaller domain gap leads to better attribution of objects, since the models are also trained with ImageNet images.

For artistic styles, while pretrained models perform well above chance (baseline Recall@10 $\approx$ 10 / 1M =  $10^{-5}$), the attribution scores are lower, indicating a more challenging task. The performance gap in different pretrained features is smaller for style-centric models (0.17-0.23) than for object-centric models (0.19-0.55). %

Interestingly, the style descriptor model ALADIN performs worse than ViT and DINO, and CLIP, even for artistic styles. We also observe that SSCD features, aimed at copy detection, do not perform well for data attribution, where synthesized images can vary drastically. This indicates that general features trained on large-scale datasests generalize to attribution more strongly than specialized features.

\myparagraph{Finetuning models.} %
Next, we evaluate how well training on object-centric and/or artistic-style variants of our data performs on those axes. 
As shown in Figure~\ref{fig:bar_plot}, interestingly, though specializing in one aspect (object or artistic style) usually leads to stronger performance on that axis, it also usually leads to performance on the other axis. Combining the two leads to consistent gains in both.

At the Pareto frontier, the DINO model, already strong on object performance, is further strengthened when training on our object attribution dataset. Interestingly, the CLIP model is not on the Pareto frontier of pretrained models, but receives a large boost when a linear layer is trained on top with our attribution data, with all 3 variants on the frontier.
Figure~\ref{fig:main_qual} shows qualitative examples that demonstrate the improved retrieval performance for finetuned CLIP and DINO attribution models, trained on both the objects and artistic-style variants.

\myparagraph{Different types of prompts.}
We study the performance when testing on different types of prompts and report results in Figure~\ref{fig:bar_plot}. For object-centric models, attribution works well overall for GPT-prompted images, whereas it is more challenging to attribute media-prompted images which induce stylistic variations.
For example, attribution becomes more ambiguous given a query such as ``{\menlo A charcoal painting of a bird}'', since either charcoal painting or realistic bird images are reasonable attributions. 

Style-centric models face a similar challenge in both prompt types, as they generate samples with the same style but different content.
However, finetuning improves the pretrained features across categories, indicating learning on our dataset is a step toward better attribution in these settings.

\myparagraph{Out-of-domain test cases.}
Figure~\ref{fig:outdomain} compares performance between in-domain (ImageNet seen classes, BAM-FG models) and out-of-domain (ImageNet unseen classes, Artchive models) test cases. Encouragingly, learning on our dataset consistently improves in-domain and out-of-domain test cases alike. This suggests that our learned attribution can generalize to different domains.

\myparagraph{Ablation studies.}
We report the effect of different mapping functions, regularization, and augmentation strategies for finetuning in Figure~\ref{fig:ablation}. We find that regularization alleviates overfitting, linear mapping works the best, and augmentation strength has a minor effect on performance. \suppmat{More details are included in Appendix~\ref{sec:supp_ablation}}

\myparagraph{Soft influence scores.}
We calculate the soft influence score described in Section~\ref{sec:attribute_nn}. %
The green text in Figure~\ref{fig:main_qual} shows the influence scores assigned to each training image. Empirically. we find that calibrating the temperature of the softmax is necessary. Since the range of the cosine similarity is small ($[-1, 1]$), assigning influence scores with default temperature ($\tau = 1$) leads to scores with insignificant variances ($< 0.0003 \%$ for all training data). After calibration, related concepts obtain significantly higher influence scores. %
\suppmat{We provide further analysis in Appendix~\ref{sec:supp_vis_calib}}. %

\begin{figure}
    \centering
    \begin{tabular}{*{1}{c}}
        \includegraphics[width=.7\linewidth]{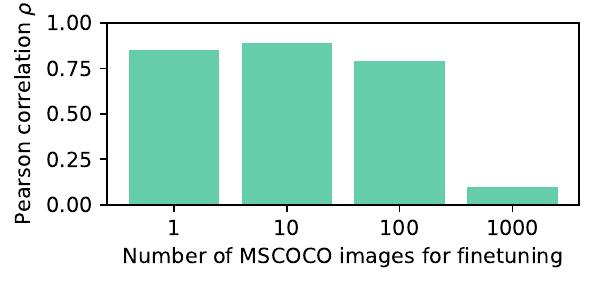}\\ \cdashline{1-1}
        \includegraphics[width=\linewidth]{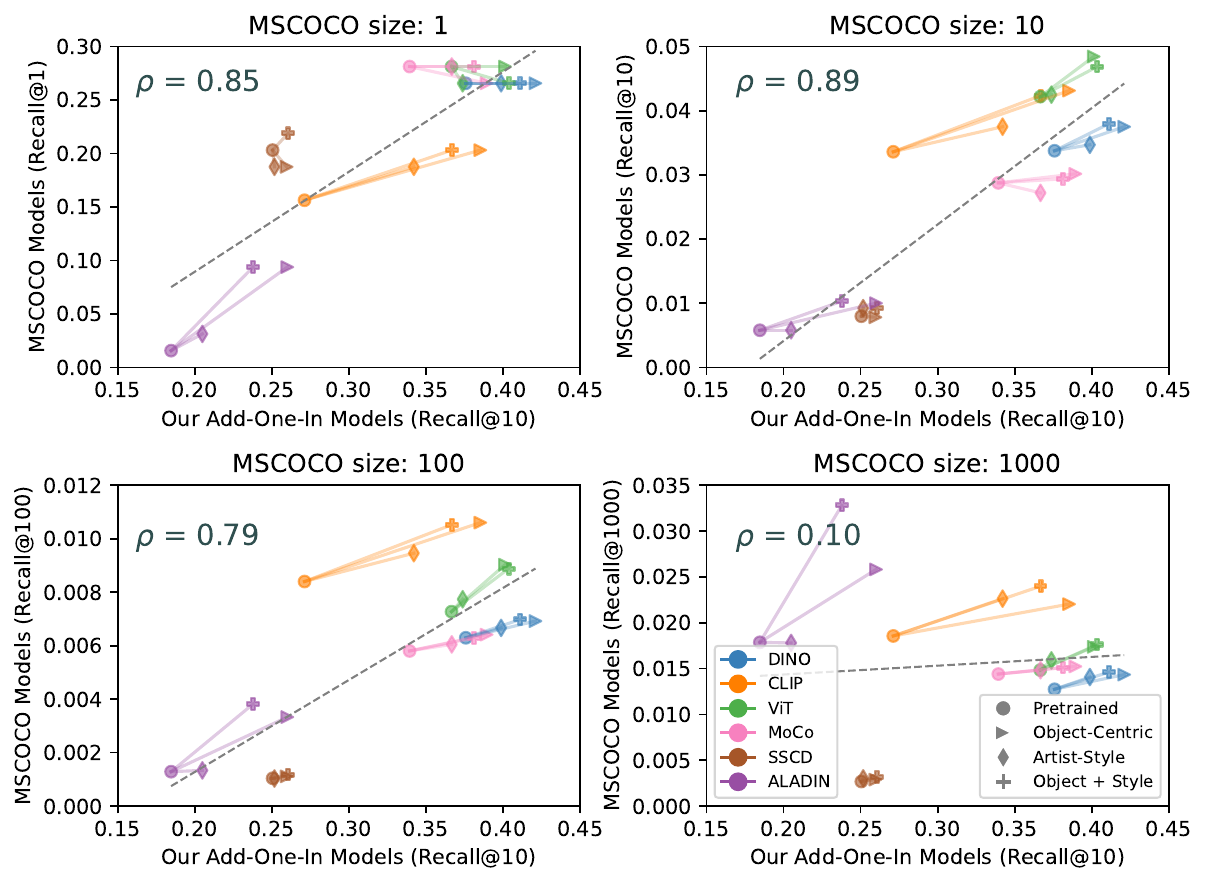}
    \end{tabular}
    \vspace{-3mm}
    \caption{\textbf{Correlation with more general attribution tasks.} \arxiv{Tuning pre-trained features on our exemplar dataset generalizes towards attribution of MSCOCO models, fine-tuned with more images.
    (Top) Pearson correlation $\rho$ with respect to the MSCOCO subset sized used for finetuning. (Bottom) The y-axes show performance on MSCOCO models (Recall@X), where X is the subset size, and x-axes show performance on our add-one-in models (Recall@10). The high correlation in 10, 100 settings indicates evaluating exemplar-based attribution correlates with a more general setting, but a gap remains with the problem of full data attribution.}}
    \vspace{-3mm}
    \label{fig:mscoco}
\end{figure}

\myparagraph{Towards general data attribution.}
\arxiv{So far, we have evaluated and learned attribution from ``add-one-in'' customized models. Does this protocol generalize to the general data attribution problem? To study this, instead of tuning towards a \textit{small} set of images representing a \textit{single} concept, we finetune with \textit{larger} sets containing \textit{multiple}, unrelated images. We draw from subsets of MSCOCO~\cite{caesar2018coco}, which provides multiple text captions for each image. We use one caption for fine-tuning, holding the rest to synthesize images for testing.
For each synthesized sample, we test whether our method can retrieve the ground truth finetuning MSCOCO images $\mathcal{X}^{+}$ from $\mathcal{X}^{+} \cup \mathcal{X}$, where $\mathcal{X}$ denotes LAION images.

We show results in Figure~\ref{fig:mscoco} for sets of size 1, 10, 100, and 1000. For most cases, (1) evaluation performance on our add-one-in models correlates with those on MSCOCO models, and (2) features tuned on our dataset (object-centric or both object+style) also improve attribution on MSCOCO models. This indicates that encouragingly, exemplar-based attribution is able to extrapolate to a more general setting. However, when there are more images (e.g., 1000), the correlation with our benchmark performance is less prominent. This suggests a gap remains between attributing exemplar models and assessing general attribution. Also, we note that plenty of headroom remains for improving attributing MSCOCO models. Additional details are in Appendix~\ref{sec:supp_mscoco}.}

\myparagraph{Qualitative results on Stable Diffusion}
We apply our finetuned features to attribute Stable Diffusion images, collected from DiffusionDB~\cite{wangDiffusionDBLargescalePrompt2022}. We run attribution on the full set of LAION-400M. Results are shown in Figure~\ref{fig:sd_nn}. Our finetuned features assign high attribution scores to related concepts (e.g., characters in console games, paintings of \emph{Last Supper}). We also note that LAION-400M contains more images with a similar concept, and the attribution scores are shared more evenly across such images.

\section{Discussion and Limitations}

Large visual generative models have not only captured the imagination of the public, but also have spawned new startups, products, and ecosystems. As such, the sourcing of training images
has brought forth important social and economic issues. A method that fairly attributes the training images opens potential possibilities where creators can be incentivized and rewarded for providing data.
\arxiv{There are good initial attempts to address data attribution~\cite{stableattribution}, and our work fills the gap by developing benchmarks to make a step toward understanding the training process and validating future attribution methods.}

\myparagraph{Limitations.}
While our method analyzes full images, tackling attribution in a compositional manner remains an open challenge. 
Even for full images, it is difficult to customize models on \textit{arbitrary} images, for example, images on LAION with unusual accompanying text prompts.
Additionally, images from pre-training, rather than just the exemplar, also exert influence, resulting in label noise. However, as customization guarantees the exemplar influence is ``upweighted'', our dataset is useful for attribution in the aggregate.

\camready{
While we train for attribution through customization, there is a domain gap to the ultimate goal of attribution for large-scale training, such as the distribution of exemplar images and selected prompts.
Additionally, there is a tradeoff between balancing exemplar influence while retaining sample diversity, and we follow best practices from Custom Diffusion~\cite{kumari2022customdiffusion}.
}
While many open challenges are left to explore in data attribution, we provide a meaningful first step toward benchmarking and understanding this area. %

{
\myparagraph{Acknowledgments.}
\noindent We thank Aaron Hertzmann for reading over an earlier draft and for insightful feedback. We thank colleagues in Adobe Research, including Eli Shechtman, Oliver Wang, Nick Kolkin, Taesung Park, John Collomosse, and Sylvain Paris, along with Alex Li and Yonglong Tian for helpful discussion. We appreciate Nupur Kumari's help with Custom Diffusion training, Ruihan Gao for proof-reading the draft, Alex Li's help to extract Stable Diffusion features, and Dan Ruta for help with the BAM-FG dataset. We thank Bryan Russell for pandemic hiking and brainstorming. This work started when SYW was an Adobe intern and was supported in part by an Adobe gift and J.P. Morgan Chase Faculty Research Award. 
}

{\small
\bibliographystyle{ieee_fullname}
\bibliography{main}
}

\section*{Appendix}
\appendix

\section{Dataset Collection Details}
\label{sec:supp_data}
In Section~\ref{sec:dataset} of the main paper, we describe using Custom Diffusion to generate ``add-one-in'' models. Here, we provide additional details. We follow the hyperparameters and best practices from Kumari~\etal~\cite{kumari2022customdiffusion}. Since only prompts related to the input image are used to build the dataset, it is unnecessary to add regularization to prevent language drifts for unrelated concepts. Thus, for faster convergence, we do not train models on the regularization datasets. We train each model \arxiv{and early stop at 125 iterations, following best practices from the paper}. %

\myparagraph{Broad categories in object-centric models.}
In Section~\ref{sec:object_models} of the main paper, we describe our object-centric models, which contain instances drawn from ImageNet classes. We group the 693 ImageNet classes into 55 broad categories, where the categories are taken from a public repo\footnote{\url{https://github.com/noameshed/novelty-detection/blob/master/imagenet_categories.csv}}. Example categories are {\menlo aircraft, arachnid, armadillo, ball, bear}, etc. These broad category names are more effective for training and prompting the model for existing model personalization works~\cite{kumari2022customdiffusion,ruiz2022dreambooth}, than the fine-grained category names, which are too long, descriptive, and do not sound like natural language, e.g., ``{\menlo yellow lady's slipper}''. We include the complete list in the attached file \href{http://www.cs.cmu.edu/~dataattribution/files/imagenet_categories.txt}{\menlo imagenet\_categories.txt} and the category assignment for each ImageNet class is in \href{http://www.cs.cmu.edu/~dataattribution/files/imagenet_class_to_categories.json}{\menlo imagenet\_class\_to\_categories.json}.

\begin{table}[h]
    \centering
    \resizebox{\linewidth}{!}{
    \begin{tabular}{cccl}
    \toprule
        Split & Model & Prompt & File  \\ \midrule \midrule
         \multirow{4}{*}{Train}& \multirow{2}{*}{\shortstack{Object-\\Centric}} & GPT & \href{http://www.cs.cmu.edu/~dataattribution/files/prompts/train/object-centric/gpt}{\menlo prompts/train/object-centric/gpt/<category>.txt}  \\
         &  & Media & \href{http://www.cs.cmu.edu/~dataattribution/files/prompts/train/object-centric/media.txt}{\menlo prompts/train/object-centric/media.txt}  \\ \cdashline{2-4}
         & \multirow{3}{*}{\shortstack{Artist-\\Style}} & GPT & \href{http://www.cs.cmu.edu/~dataattribution/files/prompts/train/style-centric/gpt.txt}{\menlo prompts/train/style-centric/gpt.txt}  \\
         & & Object (Artchive) & \href{http://www.cs.cmu.edu/~dataattribution/files/prompts/train/style-centric/object-artchive.txt}{\menlo prompts/train/style-centric/object-artchive.txt} \\
         & & Object (BAM-FG) & \href{http://www.cs.cmu.edu/~dataattribution/files/prompts/train/style-centric/object-bamfg.txt}{\menlo prompts/train/style-centric/object-bamfg.txt} \\ \midrule
         \multirow{4}{*}{Test}& \multirow{2}{*}{\shortstack{Object-\\Centric}} & GPT & \href{http://www.cs.cmu.edu/~dataattribution/files/prompts/test/object-centric/gpt}{\menlo prompts/test/object-centric/gpt/<category>.txt}  \\
         &  & Media & \href{http://www.cs.cmu.edu/~dataattribution/files/prompts/test/object-centric/media.txt}{\menlo prompts/test/object-centric/media.txt}  \\ \cdashline{2-4}
         & \multirow{3}{*}{\shortstack{Artist-\\Style}} & GPT & \href{http://www.cs.cmu.edu/~dataattribution/files/prompts/test/style-centric/gpt.txt}{\menlo prompts/test/style-centric/gpt.txt}  \\
         & & Object (Artchive) & \href{http://www.cs.cmu.edu/~dataattribution/files/prompts/test/style-centric/object-artchive.txt}{\menlo prompts/test/style-centric/object-artchive.txt} \\
         & & Object (BAM-FG) & \href{http://www.cs.cmu.edu/~dataattribution/files/prompts/test/style-centric/object-bamfg.txt}{\menlo prompts/test/style-centric/object-bamfg.txt} \\    
         \bottomrule
    \end{tabular}}
    \caption{\textbf{Prompts used in different splits.} In this document, we provide additional information regarding our prompt collection. Please see the included files for our prompts used for training and testing. Each model type and prompting type is separated into different files. {\menlo <category>} stands for each broad category used in object-centric models.}
    \label{tab:filepaths}
\end{table}

\myparagraph{More details of prompting for object-centric models.}
We create 25 ChatGPT prompts for each category with the following query:

\begin{quote}
    Provide 25 diverse image captions depicting images containing $<$category$>$, where the word ``$<$category$>$'' is in each caption as a subject. Each caption should be applicable to depict images containing any kinds of $<$category$>$ in general, without explicitly mentioning any specific $<$category$>$. Each caption should be suitable to generate realistic images using a large-scale text-to-image generative model.
\end{quote}

\noindent At times, the generated captions are repetitive or too specific, so we iterate through the querying process and manually select 25 suitable captions for each category. It is challenging to obtain over 25 suitable captions, as repetitions are likely. For each occurrence of the word {\menlo <category>}, we replace it with {\menlo $\text{V}^*$ <category>}, as the Custom Diffusion framework uses the $\text{V}^*$ token to refer to the tuned concept in the exemplar image(s). We also procedurally generate 50 prompts to introduce stylistic variations, where each prompt is of the form ``{\menlo A <medium> of $\text{V}^{*}$ <category>.}'' The 50 mediums are selected from a public repo\footnote{\url{https://github.com/pharmapsychotic/clip-interrogator/blob/main/clip_interrogator/data/mediums.txt}}.

\myparagraph{More details of prompting for artistic-style models.}
We create 50 prompts to generate painting-like captions through ChatGPT with the following query:

\begin{quote}
    Provide 50 image captions that are suitable for paintings.
\end{quote}

\noindent We iterate through the querying process and manually select 50 suitable captions. We add ``{\menlo in the style of $\text{V}^*$ art}'' at the end of each caption.

We also procedurally generate 40 prompts to introduce object variations, where each prompt is of the form ``{\menlo A picture of <object> in the style of $\text{V}^{*}$ art}'' for BAM-FG models and ``{\menlo A painting of <object> in the style of $\text{V}^{*}$ art}.'' We select 40 objects from a collection of 50 obtained by ChatGPT using the prompt:

\begin{quote}
    Provide 50 objects that can appear on a design, poster, or painting.
\end{quote}

\noindent We provide attached files with all of our prompts, listed in Table~\ref{tab:filepaths}.

\myparagraph{Prompt files.}
We use different prompts for training and testing. We include full lists of prompts for our dataset creation in the files specified in Table~\ref{tab:filepaths}. We use different prompts during training and testing, to facilitate an out-of-distribution test set.

\myparagraph{Dataset visualization.}
We visualize a subset of our dataset in the attached webpages, where we show the training exemplars, along with the samples synthesized using different prompting schemes. We show object-centric models in \href{http://www.cs.cmu.edu/~dataattribution/files/dataset_vis/object_centric.html}{\menlo dataset\_vis/object\_centric.html} and artistic-style models in \href{http://www.cs.cmu.edu/~dataattribution/files/dataset_vis/style_centric.html}{\menlo dataset\_vis/style\_centric.html}.

\section{Additional Analysis}

\definecolor{gggreen}{RGB}{50, 128, 30}
\begin{figure}
    \centering
    \includegraphics[width=\linewidth]{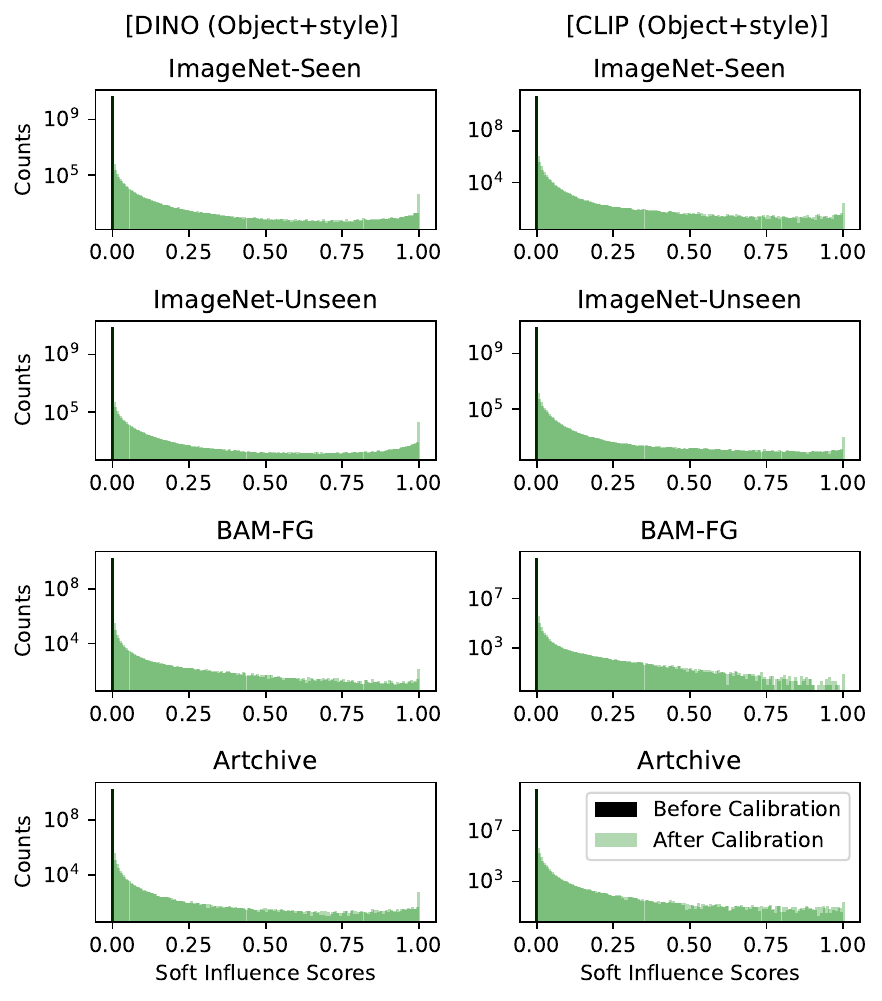}
    \caption{\textbf{Distribution of influence scores.} We collect the influence scores assigned to the training points from each query and visualize the distribution of scores. We show the results of our tuned DINO features (Left) and our tuned CLIP features (Right). Each row represents a test case. We compare the distribution before and after calibration, colored in \textbf{black} and \textbf{{\color{gggreen} green}}, respectively. The uncalibrated softmax gives small values close to 0, and calibration produces more meaningful influence scores.}
    \label{fig:softmax}
\end{figure}

\subsection{Visualizing the calibrated influence scores.}
\label{sec:supp_vis_calib}
In the main paper, we formulate the soft influence score calibration in Section~\ref{sec:attribute_nn} and discuss the results in Section~\ref{sec:exp}. Here, we provide additional analysis by visualizing the difference between calibrated influence score and a plain softmax applied on the feature similarities. Figure~\ref{fig:softmax} shows the distribution of influence scores before and after calibration. Since the range of the cosine similarity is small ($[ -1, 1 ]$), applying just softmax on the similarities leads to evenly spread influence scores across the training data. This indicates the need to find an appropriate temperature to give a more distinctive influence score. In Figure~\ref{fig:softmax}, after calibration, we can assign higher influence scores to the top-attributed training images.

\subsection{Ablation Studies}
\label{sec:supp_ablation}
In Section~\ref{sec:exp} of the main text, we report the effect of different mapping functions, regularization, and augmentation strategies for finetuning in Figure~\ref{fig:ablation}. In this section, we discuss the ablation in more detail. %

\myparagraph{Regularization.} First, learning the linear mappings without regularization leads to significant overfitting. Since our regularization encourages linear mapping to be a rigid transform, the learned mapping with regularization preserves pretrained features' properties, resulting in better generalization.

\myparagraph{Mapping functions.} Next, we compare two other choices of mapping functions with increasing capacity.
Channel-wise multiplication (\textbf{Channel}) can only scale the existing features and is not enough to improve the performance. On the other hand, we try increasing the capacity using 2-layer MLP (\textbf{MLP}), resembling the projection head in MoCo v3~\cite{chen2021empirical}. However, this leads to overfitting, resulting in a much worse generalization on the test set. This validates that our final setting of a well-regularized linear mapping function (\textbf{Linear}) is a sweet spot, with enough complexity to improve performance, without suffering from overfitting.

\myparagraph{Data augmentations.} Finally, we compare performance when trained with mild and strong augmentations. Our mild augmentation consists of random horizontal flips, resizing, and crops. Our strong augmentation follows DINO~\cite{caron2021emerging} (random flips, resizing, crops, color jittering,
Gaussian blur and solarization, with multi-crops removed). We find that two types of augmentation schemes yield similar performance, where applying mild augmentation is slightly favorable. This indicates that applying strong augmentation is not necessary, as we are taking advantage of a pretrained feature extractor and learning light mapping function on top. Hence, we use mild augmentation throughout other experiments.

\begin{figure}
    \centering
    \includegraphics[width=\linewidth]{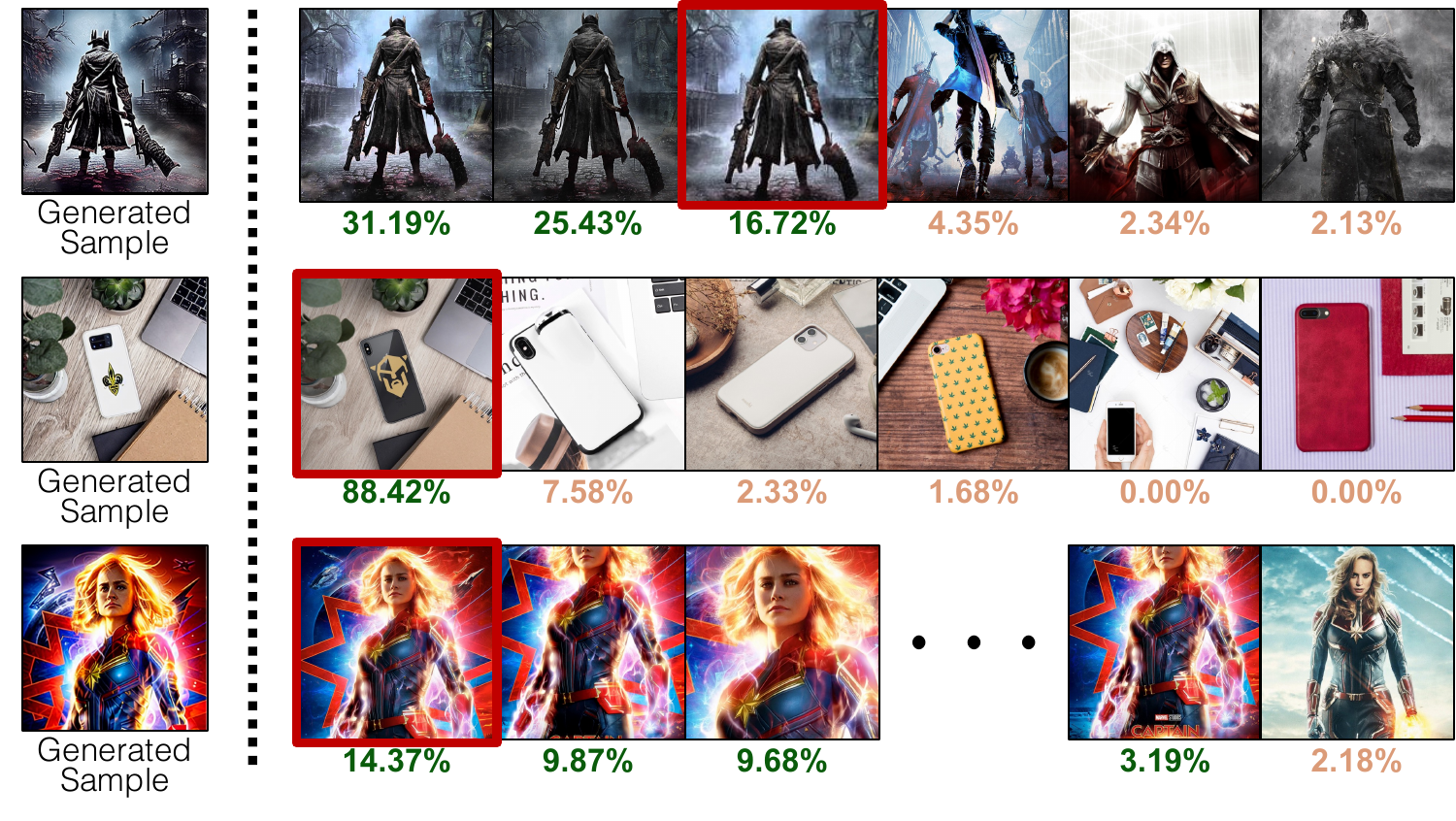}
    \caption{\textbf{Copy detection results.} We investigate our data attribution method when a synthesized image query is a ``duplicate'' of the training data. We take duplicates found in Somepalli~\etal~\cite{somepalli2022diffusion}, where the images in the red box are the training image matches reported by the authors. We observe that there exist multiple training images from LAION that resemble the query, and the attribution score dropoff ({\color{gggreen} green} $\rightarrow$ {\color{orange} orange}) is significant between similar images and other images.} 
    \label{fig:supp_copy}
\end{figure}

\subsection{Additional Stable Diffusion results.}
\myparagraph{Copy detection for Stable Diffusion.}
We investigate the extreme case where the generated sample is a memorization of a training point. We test with several pairs of memorized samples and corresponding training image matches reported in Somepalli~\etal~\cite{somepalli2022diffusion}. We add the training image matches into the 1M random LAION subset used in Section 5 of our main pair, and for each memorized sample, we assign attribution scores to the augmented set of training images. Results are shown in Figure~\ref{fig:supp_copy}. Our method assigns high attribution scores to the matched training images. We also find that there exist multiple training images similar to the memorized samples, and that there is a significant attribution score dropoff between similar images and the other training images.

\myparagraph{Stable Diffusion attribution.}
In Section 5 of the main paper, we show qualitative results on attributing images generated by Stable Diffusion. Here we show additional samples in Figure~\ref{fig:supp_sd_nn}. In practice, for each query, we retrieve the top 10,000 images from LAION-400M and assess attribution scores. In most cases, we find that the attribution scores are already zero for the lower ranked images in this subset.

\subsection{Additional Custom diffusion results.}
\label{sec:supp_additional_metrics}
\begin{figure*}
    \centering
    \begin{tabular}{cc}
        \includegraphics[width=0.42\linewidth]{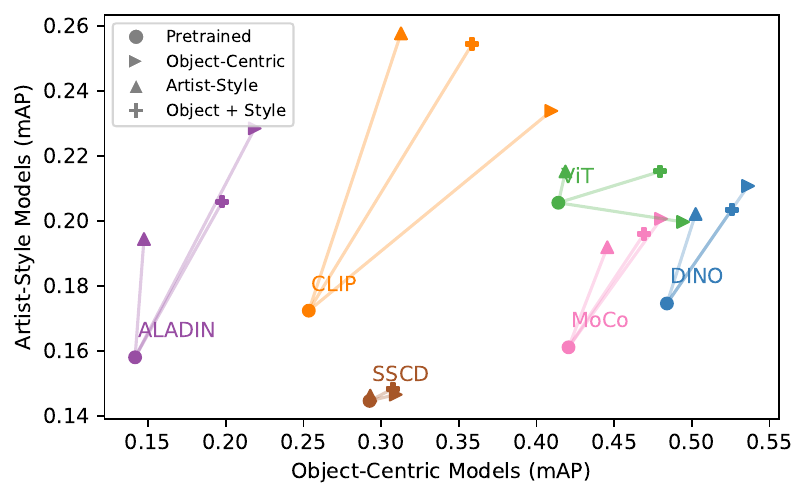} & 
        \includegraphics[width=0.5\linewidth]{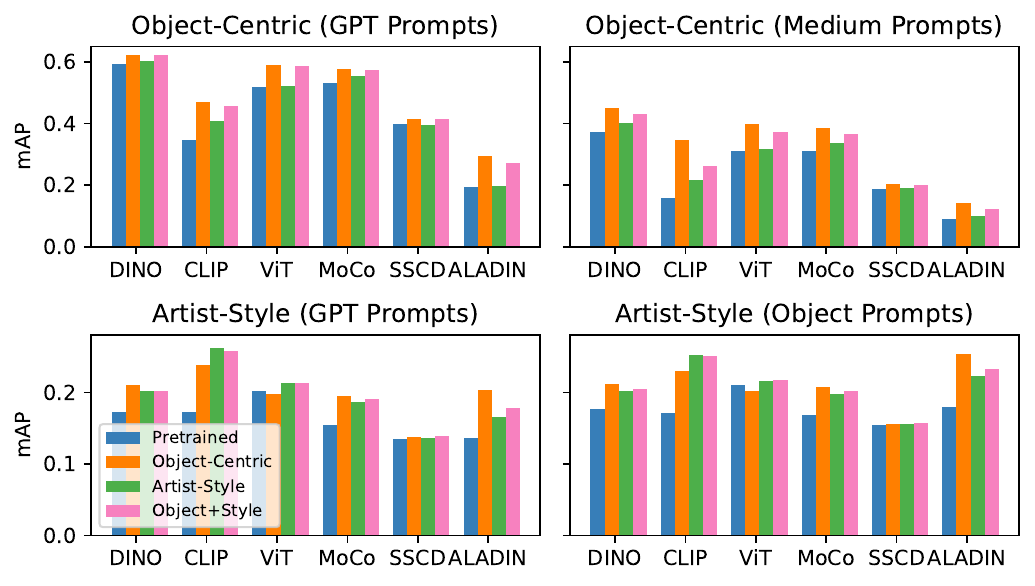} \\ \midrule
        \includegraphics[width=0.42\linewidth]{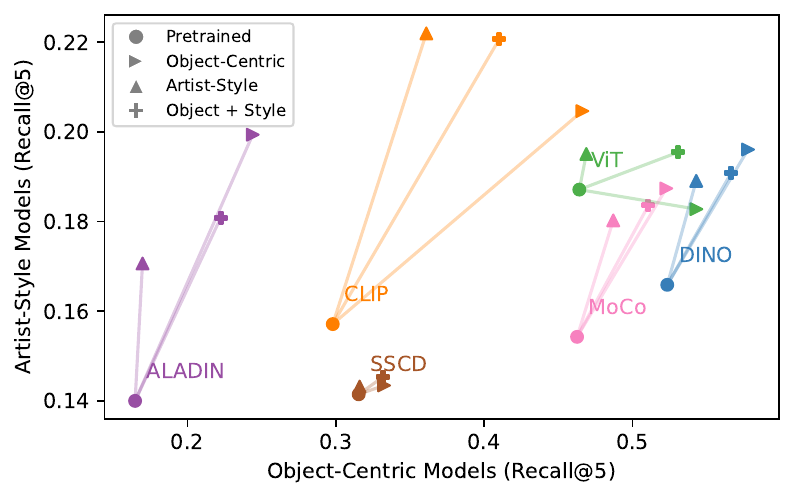} & 
        \includegraphics[width=0.5\linewidth]{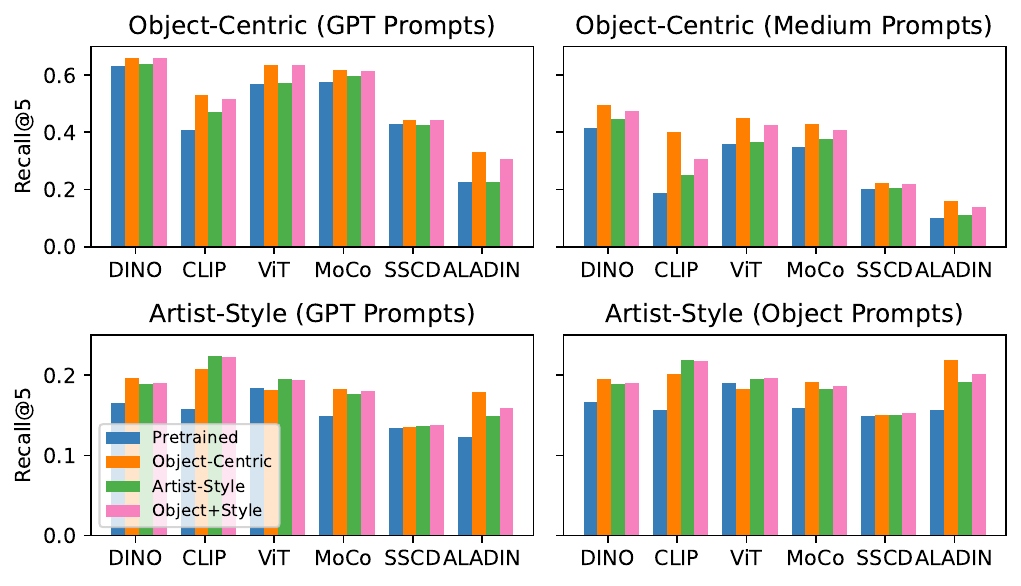} \\ \midrule
        \includegraphics[width=0.42\linewidth]{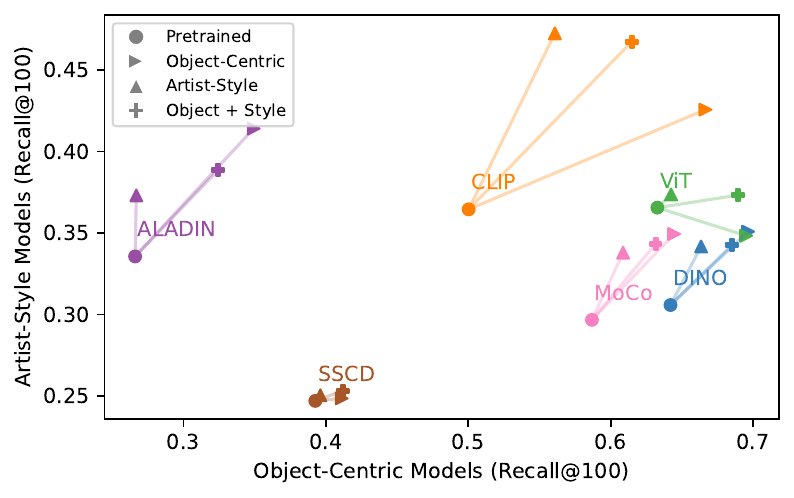} & 
        \includegraphics[width=0.5\linewidth]{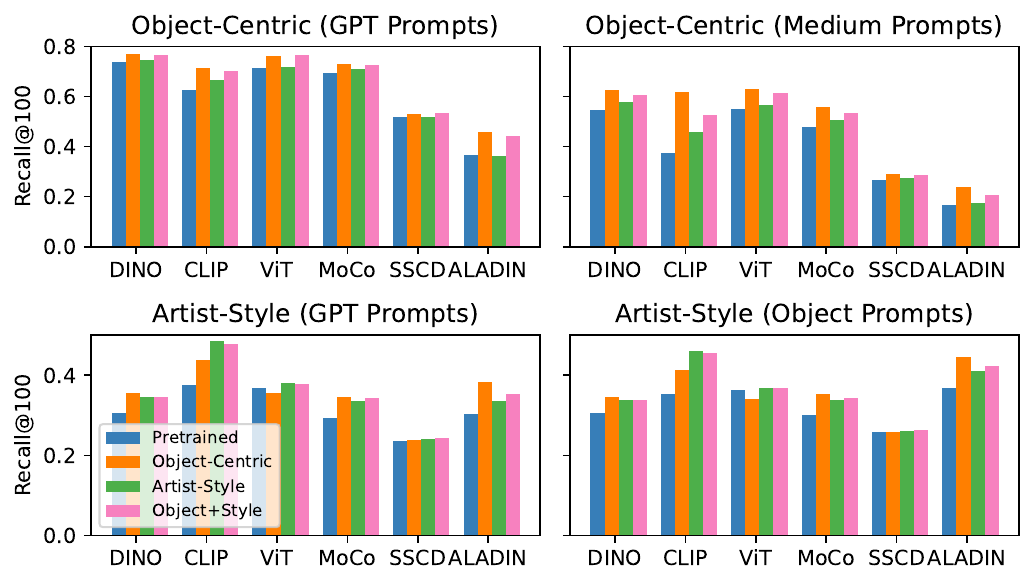} \\
    \end{tabular}
    \caption{\textbf{Additional metrics.} We show the same visualization as in Figure~\ref{fig:bar_plot} in the main text with additional metrics (top: mAP, middle: Recall@5, bottom: Recall@100). We find that all metrics give similar trends and rankings across different methods and test cases.}
    \label{fig:sup_bar_plot}
\end{figure*}

\myparagraph{Additional metrics.}
In Section 5 of the main paper, we use Recall@10 for the analysis. Here we include more metrics (Recall@1, Recall@100, mAP) and report the evaluation of the in-domain and out-of-domain test cases in Table~\ref{tab:indomain} and Table~\ref{tab:outdomain}, respectively. We also show that the general trend and rankings across different methods and feature spaces hold the same across different metrics in Figure~\ref{fig:sup_bar_plot}.

\begin{figure*}
    \centering
    \includegraphics[width=\linewidth]{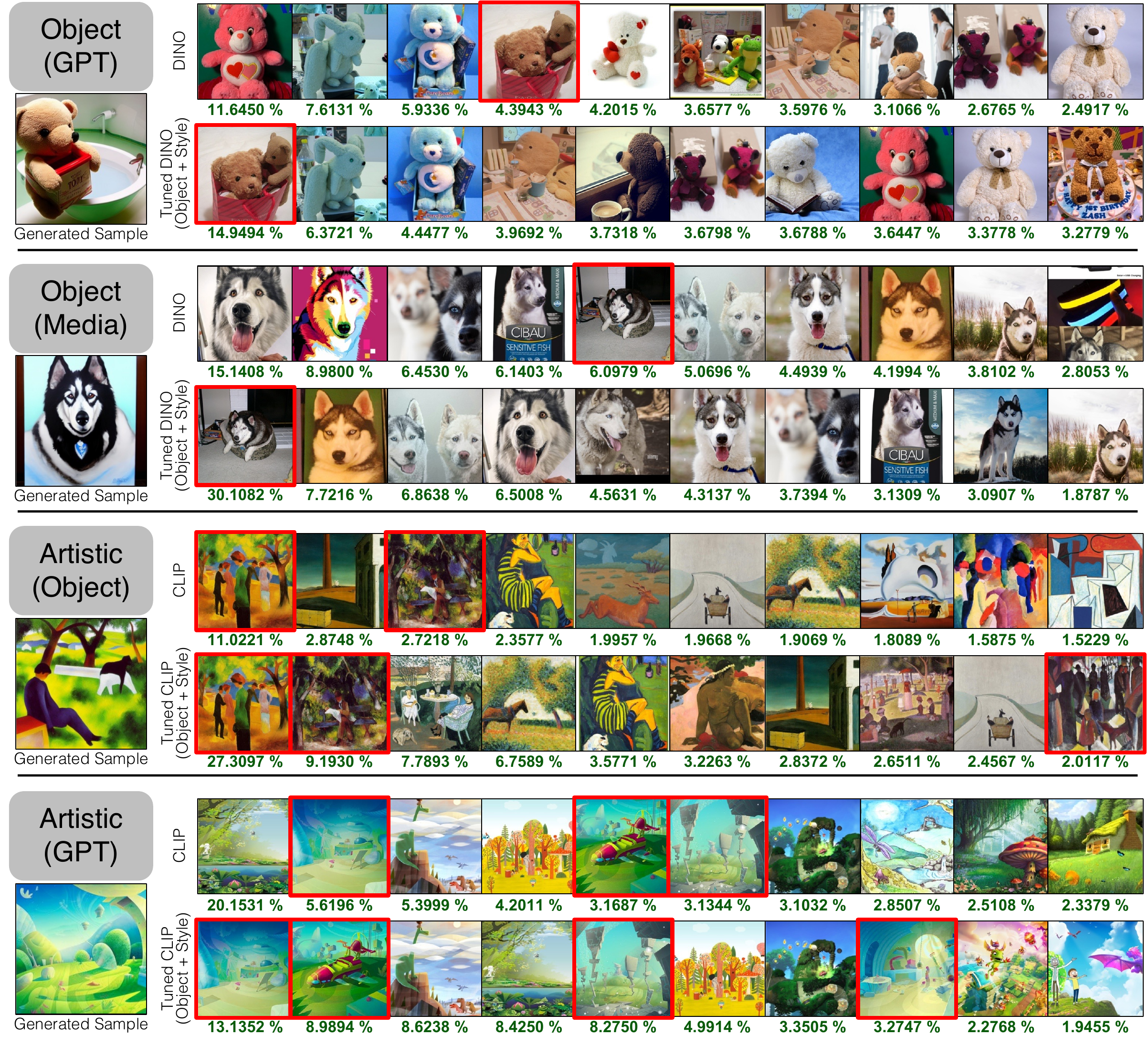}
    \caption{\textbf{Additional results on qualitative comparisons.} We show results in addition to Figure~\ref{fig:main_qual} of the main paper. Here we show one query from each model type (Object-Centric, Artistic-Style) and each prompting type (GPT-generated, procedurally generated prompt). We show comparisons between pretrained features and features finetuned on our Object+Style attribution dataset. For object-centric models (top two rows), we show results using DINO as the base encoder, and for artistic-style models (bottom two rows), we show results using CLIP as the base encoder. {\bf {\color{gggreen} Green}} values are calibrated influence percentage scores. We find that our fine-tuned attribution method improves the ranking and influence score of the exemplar training images (red-boxed images). }
    \label{fig:supp_attr}
\end{figure*}

\myparagraph{Qualitative results on Custom Diffusion.}
In Section~\ref{sec:exp} of the main paper, we have included qualitative comparisons between pretrained and fine-tuned features for attributing our Custom Diffusion dataset. We provide more qualitative results in Figure~\ref{fig:supp_attr}.

\subsection{Additional Baselines.}

We evaluate additional baselines -- using Stable Diffusion as the feature extractor, and testing whether customization is needed.

\myparagraph{Attribution with Stable Diffusion features.}
\arxiv{In the main paper, we use feature spaces that only analyze the images. Here, we test an additional baseline, using Stable Diffusion features, which also incorporates the image captions. We feed images into the pre-trained Stable Diffusion model, along with captions, to collect features. We use the provided captions for LAION images, the input prompts for synthetic images (with {\menlo V$^{*}$} token removed), and apply captioning model BLIP~\cite{li2022blip} for the exemplar images (as is standard practice for captionless images~\cite{parmar2023zero}). Following Li~\etal~\cite{li2023diffusion}, we obtain the average-pooled, mid-layer, U-Net features at timestep $t = 100$ (model is trained with 1000-step DDPM).}

\arxiv{For object-centric models, this baseline obtains a lower score (0.279) than DINO and CLIP (0.553, 0.342) at Recall@10. For style-centric models, the baseline (0.202) obtains similar scores to DINO and CLIP (0.199, 0.201). 
Thus, Stable Diffusion features do \emph{not} outperform the features evaluated in our paper. Incorporating text to improve attribution is an intriguing future research avenue, and our dataset can spur future work in this area.}

\myparagraph{Learning attribution without ``add-one-in'' training.}
\arxiv{We study whether ``add-one-in'' customization is necessary for providing useful training signal for attribution. To investigate this, we train on a dataset where all images are generated \emph{without} finetuning Stable Diffusion.
Using the same prompts in our dataset, we generate image variations to create real-synthetic attribution pairs, but here {\menlo V$^{*}$ <noun>} is swapped with the BLIP-generated caption from the real image. We take 593 real images each from object-centric and style-centric exemplars and synthesize ~70k images in total. We fine-tune DINO and CLIP features on this dataset and compare them against those tuned on a subset of our dataset of the same size. Tuning on our dataset leads to improvements on vanilla features -- $0.376\rightarrow 0.406$ for DINO, but $0.393$ for the baseline (for Recall@10). For CLIP, our method improves results to $0.271\rightarrow 0.363$, whereas the baseline achieves $0.359$. Thus, we find that tuning on our dataset (learned from Custom Diffusion), outperforms the baseline. }

\section{Implementation Details}

\subsection{Training details for feature mappers}
\label{sec:supp_mapper_train_detail}
We initialize the linear mapper with an identity weight matrix and zero bias. We use Adam optimizer~\cite{kingma2014adam} with $\beta_1 = 0.9, \beta_2 = 0.999$. We apply an initial learning rate $10^{-3}$ and decay the learning rate with a cosine schedule~\cite{loshchilov2016sgdr}. We set the batch size to 1024 for all cases except for ViT and ALADIN. We use 512 due to memory constraints.

We train each model for 100 epochs, where each epoch involves sampling an attribution pair from each model exactly \emph{once}. During training, we sample a minibatch of $B$ different models. Within each model, we randomly choose the training image and prompting type, and we then randomly choose the synthesized image given the prompting type.

We use the same input size and image normalization as each base feature encoder. During training, we randomly flip, resize, and crop to the input size. During testing, we apply center cropping to the input size.

\begin{figure}
    \centering
    \includegraphics[width=\linewidth]{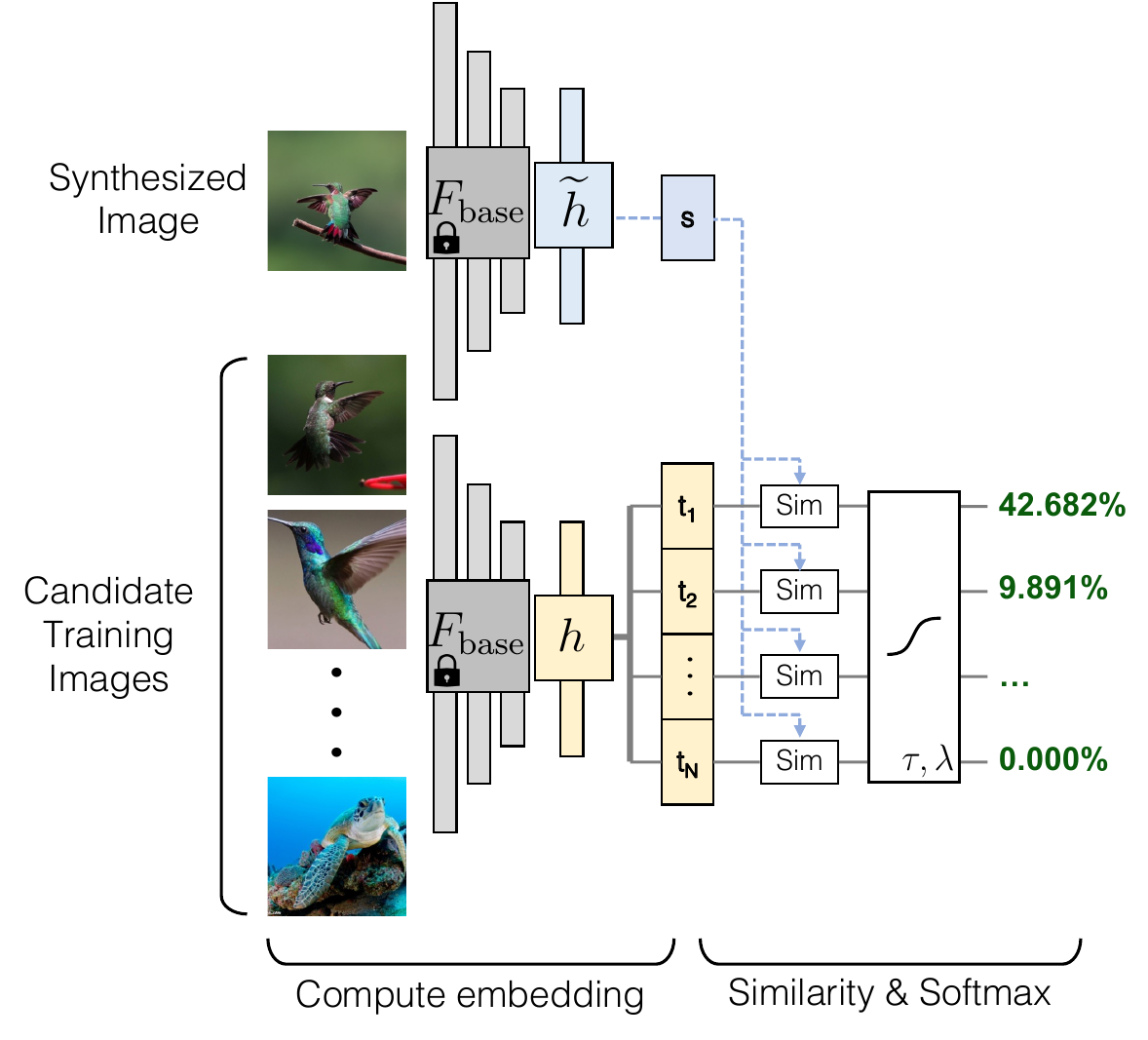}
    \caption{\textbf{Inference pipeline.}  We compute the similarity between the synthesized and training images, using a base feature extractor and our learned embedding. The training procedure is illustrated in Figure 3 in the main paper. Taking a thresholded softmax with calibrated temperature $\tau$ and threshold $\lambda$ over similarities produces influence scores.} 
    \label{fig:inference}
\end{figure}

\subsection{Details for fine-tuning on MSCOCO}
\label{sec:supp_mscoco}
\arxiv{We finetune Stable Diffusion models on random subset of MSCOCO of size 1, 10, 100, or 1000. For each subset size, we sample 4 random subsets to train 4 models and average the results. We adopt similar finetuning technique as Custom Diffusion, where we only tune the cross-attention weights, but here we use MSCOCO captions to train the models instead of associating a concept to a {\menlo $V^{*}$} token. We adopt the same hyperparameters used in Custom Diffusion, and for MSCOCO with sizes 1, 10, 100, and 1000, our training iterations are 125, 250, 500, 2000, and 20000, respectively. We synthesize 4 samples per hold-out caption to generate our queries, except for models trained with 1000 MSCOCO images, we synthesize 1 sample per caption given the large number of captions available in this test case.}
\subsection{Soft influence score implementation}
\label{sec:supp_soft}
\arxiv{In Section~\ref{sec:attribute_nn} of the main paper, we describe how we convert and calibrate feature similarities to a percentage assignment $\hat{P}_{\tau,\lambda}$ via Equation~\ref{eqn:calib_score_func}, repeated below. We provide additional details on how we optimize for $\tau$ and $\lambda$.}

\begin{equation}
    \begin{aligned}
        \hat{P}_{\tau,\lambda}({\bf x} |\synthpt; \mathcal{X}^{+}  \cup \mathcal{X}) =  \frac{ \text{ReLU} \big( \exp[ \sfrac{(s - s_{(0)})}{\tau}] - \lambda \big)}{ \sum_j \text{ReLU}\big( \exp [ \sfrac{(s_{(j)} - s_{(0)})}{\tau} ] - \lambda \big) }
    \end{aligned}
    \label{eqn:supp_calib_score_func}
\end{equation}

\noindent \arxiv{Recall that $\synthpt$ is a synthetic content query, ${\bf x \in \mathcal{X}^{+} \cup \mathcal{X}}$ denotes images in the exemplar-augmented training set $\mathcal{X}^{+} \cup \mathcal{X}$. We denote $s$ as the feature similarity between $\synthpt$ and ${\bf x}$, and we write them in descending order: $s_{(0)}\geq s_{(1)}\geq \dots\geq s_{(|\mathcal{X}^+ \cup \mathcal{X}|)}$. }

\arxiv{As $\mathcal{X}^{+}  \cup \mathcal{X}$ is a large collection of images (original dataset $\mathcal{X}$ is 1M images), an exhaustive calculation would be prohibitively costly. Hence, during calibration, we keep the top $R=100,000$ similarity scores individually, corresponding to the top $10\%$, along with the ground truth if needed (which is typically in the top-$10\%$ matches already). We discard the remaining $|\mathcal{X}^+ \cup \mathcal{X}|-R$ points since these terms will likely be compressed to zero by ReLU. During optimization, we find that directly using ReLU leads to instability. To resolve this, we approximate the ReLU using a smooth softplus function during training.}

\arxiv{During training, we set the learning rate for $\tau$ and $\lambda$ to be $0.5$ and $0.0005$, respectively, since we find that a high learning rate for the threshold $\lambda$ will lead to instability. $\tau$ and $\lambda$ are initialized as $1$ and $0$, respectively. We set the batch size to be $4096$ and train with $200$ steps. We also find that setting the softplus $\beta$ parameter to be $100$ helps stabilize training.}

\arxiv{For the retrieval task, we purposely train \emph{without} LAION to avoid learning a LAION classifier. We note that tune the influence scores on LAION, but we do \emph{not} change the learned feature similarities.}

\begin{figure*}
    \centering
    \includegraphics[width=\linewidth]{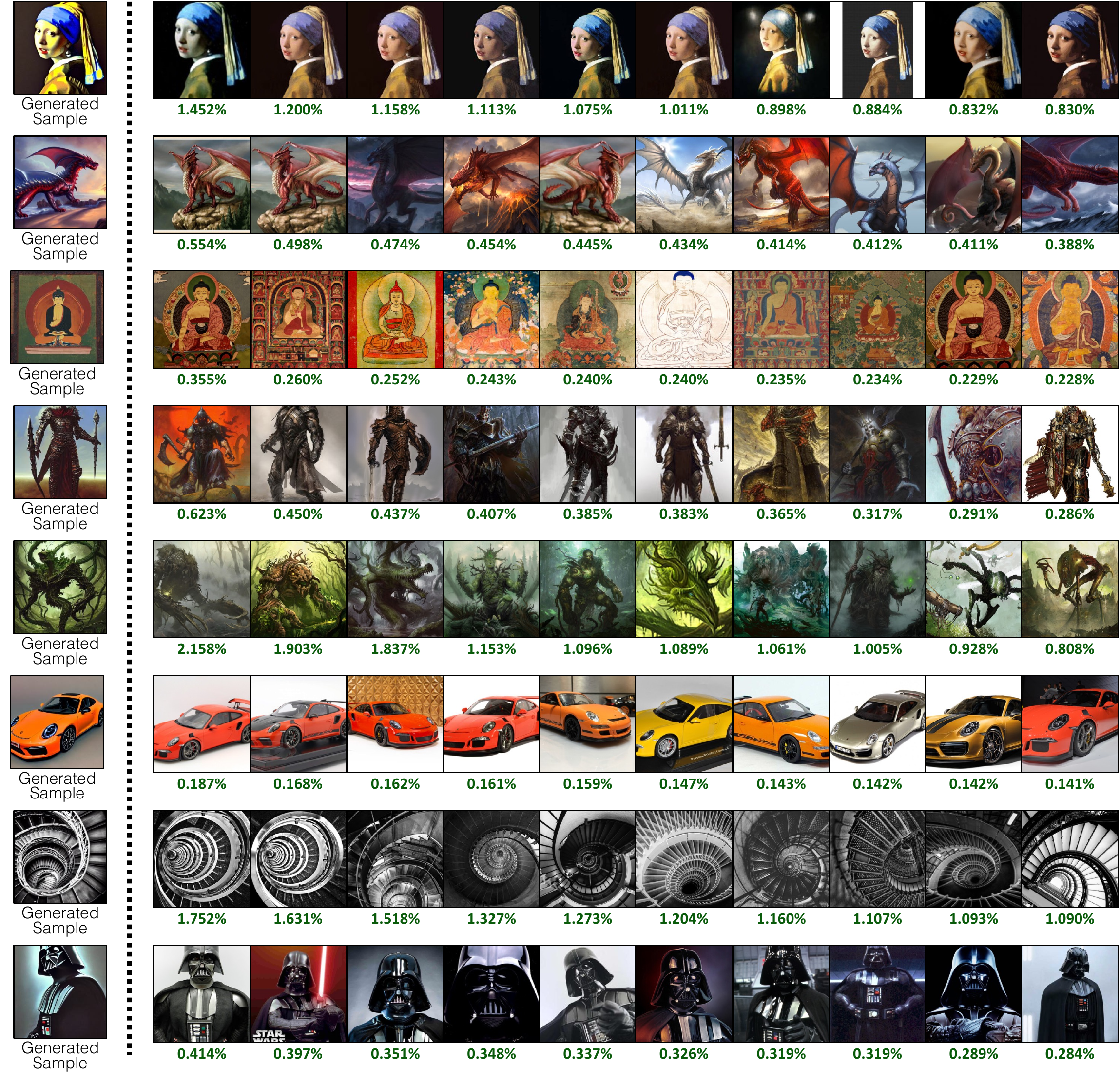}
    \caption{\textbf{Additional results on attributing Stable Diffusion Images.} We show results in addition to Figure~\ref{fig:sd_nn} of the main paper. We run our influence score prediction function with CLIP, tuned on our Object+Style attribution datasets. In each row, we show a generated sample query (Left), and the top attributed training images from LAION-400M (Right). {\bf {\color{gggreen} Green}} values are calibrated influence percentage scores.}
    \label{fig:supp_sd_nn} 
\end{figure*}

\begin{table*}[]
\resizebox{\linewidth}{!}{
\begin{tabular}{llcccccccccccccccc}
\toprule
\multicolumn{2}{l}{Source}                           & \multicolumn{8}{c}{ImageNet-Seen}                             & \multicolumn{8}{c}{BAM-FG}                                    \\ \cmidrule(lr){1-2}\cmidrule(lr){3-10} \cmidrule(lr){11-18}
\multicolumn{2}{l}{Prompts}                          & \multicolumn{4}{c}{GPT}       & \multicolumn{4}{c}{Medium}    & \multicolumn{4}{c}{GPT}       & \multicolumn{4}{c}{Object}    \\ \cmidrule(lr){1-2}\cmidrule(lr){3-6}\cmidrule(lr){7-10} \cmidrule(lr){11-14}\cmidrule(lr){15-18}
$F_\text{base}$         & Method                     & R@5   & R@10  & R@100 & mAP   & R@5   & R@10  & R@100 & mAP   & R@5   & R@10  & R@100 & mAP   & R@5   & R@10  & R@100 & mAP   \\ \midrule
\multirow{8}{*}{CLIP}   & Pretrained                 & 0.236 & 0.277 & 0.437 & 0.195 & 0.137 & 0.160 & 0.274 & 0.118 & 0.129 & 0.170 & 0.310 & 0.148 & 0.174 & 0.225 & 0.374 & 0.200 \\ \cdashline{2-18} 
                        & Object                     & 0.350 & 0.397 & 0.561 & 0.298 & 0.293 & 0.336 & 0.503 & 0.249 & 0.166 & 0.218 & 0.364 & 0.195 & 0.216 & 0.275 & 0.431 & 0.252 \\
                        & Style                      & 0.277 & 0.317 & 0.478 & 0.232 & 0.176 & 0.204 & 0.346 & 0.153 & 0.185 & 0.242 & 0.405 & 0.218 & 0.235 & 0.302 & 0.472 & 0.278 \\ \cdashline{2-18} 
                        & Object+Style (No reg.)     & 0.223 & 0.266 & 0.453 & 0.180 & 0.137 & 0.169 & 0.334 & 0.110 & 0.107 & 0.148 & 0.294 & 0.124 & 0.141 & 0.187 & 0.352 & 0.160 \\ \cdashline{2-18} 
                        & Object+Style (Channel)     & 0.238 & 0.277 & 0.431 & 0.197 & 0.141 & 0.163 & 0.278 & 0.120 & 0.136 & 0.181 & 0.328 & 0.158 & 0.186 & 0.242 & 0.406 & 0.217 \\
                        & Object+Style (MLP)         & 0.133 & 0.174 & 0.380 & 0.100 & 0.082 & 0.113 & 0.295 & 0.064 & 0.062 & 0.091 & 0.226 & 0.074 & 0.079 & 0.114 & 0.274 & 0.093 \\
                        & Object+Style (Strong Aug.) & 0.326 & 0.373 & 0.539 & 0.276 & 0.222 & 0.260 & 0.434 & 0.189 & 0.176 & 0.231 & 0.390 & 0.206 & 0.224 & 0.287 & 0.453 & 0.262 \\ \cdashline{2-18} 
                        & Object+Style               & 0.329 & 0.376 & 0.540 & 0.280 & 0.222 & 0.258 & 0.428 & 0.190 & 0.186 & 0.243 & 0.408 & 0.219 & 0.236 & 0.303 & 0.474 & 0.278 \\ \midrule
\multirow{8}{*}{DINO}   & Pretrained                 & 0.433 & 0.467 & 0.579 & 0.393 & 0.288 & 0.321 & 0.428 & 0.255 & 0.148 & 0.184 & 0.293 & 0.163 & 0.193 & 0.239 & 0.360 & 0.219 \\ \cdashline{2-18} 
                        & Object                     & 0.479 & 0.517 & 0.628 & 0.437 & 0.368 & 0.399 & 0.511 & 0.325 & 0.168 & 0.208 & 0.324 & 0.188 & 0.216 & 0.267 & 0.391 & 0.247 \\
                        & Style                      & 0.443 & 0.476 & 0.590 & 0.402 & 0.315 & 0.348 & 0.461 & 0.278 & 0.164 & 0.204 & 0.318 & 0.183 & 0.210 & 0.257 & 0.381 & 0.237 \\ \cdashline{2-18} 
                        & Object+Style (No reg.)     & 0.324 & 0.368 & 0.532 & 0.275 & 0.211 & 0.245 & 0.385 & 0.176 & 0.056 & 0.078 & 0.176 & 0.060 & 0.077 & 0.103 & 0.220 & 0.082 \\ \cdashline{2-18} 
                        & Object+Style (Channel)     & 0.377 & 0.409 & 0.522 & 0.337 & 0.233 & 0.259 & 0.367 & 0.204 & 0.109 & 0.137 & 0.232 & 0.118 & 0.148 & 0.182 & 0.291 & 0.163 \\
                        & Object+Style (MLP)         & 0.171 & 0.220 & 0.434 & 0.132 & 0.099 & 0.133 & 0.309 & 0.077 & 0.024 & 0.036 & 0.120 & 0.027 & 0.030 & 0.046 & 0.144 & 0.034 \\
                        & Object+Style (Strong Aug.) & 0.473 & 0.507 & 0.621 & 0.431 & 0.347 & 0.379 & 0.491 & 0.307 & 0.162 & 0.201 & 0.315 & 0.180 & 0.208 & 0.255 & 0.379 & 0.236 \\ \cdashline{2-18} 
                        & Object+Style               & 0.475 & 0.510 & 0.623 & 0.433 & 0.351 & 0.383 & 0.496 & 0.311 & 0.165 & 0.205 & 0.320 & 0.184 & 0.212 & 0.259 & 0.383 & 0.239 \\ \midrule
\multirow{5}{*}{MoCo}   & Pretrained                 & 0.390 & 0.425 & 0.537 & 0.347 & 0.239 & 0.267 & 0.372 & 0.211 & 0.130 & 0.163 & 0.273 & 0.144 & 0.187 & 0.232 & 0.355 & 0.211 \\ \cdashline{2-18} 
                        & Object                     & 0.443 & 0.479 & 0.589 & 0.400 & 0.313 & 0.345 & 0.457 & 0.278 & 0.151 & 0.191 & 0.307 & 0.171 & 0.211 & 0.261 & 0.392 & 0.242 \\
                        & Style                      & 0.409 & 0.442 & 0.555 & 0.365 & 0.263 & 0.292 & 0.400 & 0.232 & 0.150 & 0.188 & 0.303 & 0.168 & 0.207 & 0.255 & 0.381 & 0.235 \\ \cdashline{2-18} 
                        & Object+Style (No reg.)     & 0.336 & 0.379 & 0.532 & 0.287 & 0.221 & 0.255 & 0.385 & 0.187 & 0.063 & 0.086 & 0.188 & 0.068 & 0.087 & 0.117 & 0.241 & 0.095 \\ \cdashline{2-18} 
                        & Object+Style               & 0.437 & 0.472 & 0.581 & 0.394 & 0.295 & 0.327 & 0.435 & 0.262 & 0.153 & 0.192 & 0.308 & 0.172 & 0.209 & 0.258 & 0.385 & 0.238 \\ \midrule
\multirow{5}{*}{ViT}    & Pretrained                 & 0.355 & 0.393 & 0.530 & 0.310 & 0.242 & 0.275 & 0.413 & 0.210 & 0.168 & 0.215 & 0.343 & 0.193 & 0.224 & 0.278 & 0.415 & 0.259 \\ \cdashline{2-18} 
                        & Object                     & 0.452 & 0.489 & 0.615 & 0.406 & 0.335 & 0.372 & 0.509 & 0.294 & 0.172 & 0.218 & 0.342 & 0.196 & 0.221 & 0.274 & 0.405 & 0.256 \\
                        & Style                      & 0.357 & 0.396 & 0.539 & 0.314 & 0.253 & 0.290 & 0.432 & 0.219 & 0.182 & 0.230 & 0.359 & 0.209 & 0.231 & 0.288 & 0.425 & 0.270 \\ \cdashline{2-18} 
                        & Object+Style (No reg.)     & 0.261 & 0.312 & 0.505 & 0.212 & 0.176 & 0.217 & 0.391 & 0.140 & 0.087 & 0.122 & 0.259 & 0.102 & 0.123 & 0.167 & 0.335 & 0.142 \\ \cdashline{2-18} 
                        & Object+Style               & 0.448 & 0.487 & 0.618 & 0.398 & 0.315 & 0.353 & 0.492 & 0.274 & 0.182 & 0.230 & 0.358 & 0.209 & 0.234 & 0.291 & 0.428 & 0.272 \\ \midrule
\multirow{5}{*}{ALADIN} & Pretrained                 & 0.108 & 0.125 & 0.205 & 0.090 & 0.070 & 0.080 & 0.120 & 0.062 & 0.115 & 0.155 & 0.273 & 0.140 & 0.195 & 0.259 & 0.417 & 0.236 \\ \cdashline{2-18} 
                        & Object                     & 0.189 & 0.211 & 0.297 & 0.162 & 0.115 & 0.128 & 0.184 & 0.103 & 0.154 & 0.206 & 0.340 & 0.187 & 0.253 & 0.331 & 0.495 & 0.308 \\
                        & Style                      & 0.112 & 0.131 & 0.206 & 0.095 & 0.079 & 0.088 & 0.127 & 0.070 & 0.150 & 0.200 & 0.329 & 0.181 & 0.247 & 0.322 & 0.485 & 0.300 \\ \cdashline{2-18} 
                        & Object+Style (No reg.)     & 0.101 & 0.128 & 0.254 & 0.077 & 0.063 & 0.077 & 0.151 & 0.050 & 0.088 & 0.123 & 0.256 & 0.104 & 0.134 & 0.183 & 0.355 & 0.157 \\ \cdashline{2-18} 
                        & Object+Style               & 0.172 & 0.194 & 0.281 & 0.149 & 0.103 & 0.114 & 0.164 & 0.092 & 0.152 & 0.202 & 0.332 & 0.183 & 0.249 & 0.324 & 0.487 & 0.302 \\ \midrule
\multirow{5}{*}{SSCD}   & Pretrained                 & 0.253 & 0.272 & 0.335 & 0.230 & 0.142 & 0.153 & 0.194 & 0.131 & 0.117 & 0.146 & 0.231 & 0.128 & 0.175 & 0.214 & 0.313 & 0.194 \\ \cdashline{2-18} 
                        & Object                     & 0.265 & 0.284 & 0.351 & 0.241 & 0.158 & 0.170 & 0.215 & 0.144 & 0.113 & 0.141 & 0.222 & 0.123 & 0.169 & 0.207 & 0.302 & 0.187 \\
                        & Style                      & 0.254 & 0.273 & 0.339 & 0.228 & 0.144 & 0.155 & 0.199 & 0.132 & 0.119 & 0.148 & 0.234 & 0.130 & 0.174 & 0.215 & 0.315 & 0.193 \\ \cdashline{2-18} 
                        & Object+Style (No reg.)     & 0.056 & 0.070 & 0.144 & 0.045 & 0.035 & 0.042 & 0.081 & 0.029 & 0.021 & 0.030 & 0.077 & 0.022 & 0.032 & 0.044 & 0.103 & 0.033 \\ \cdashline{2-18} 
                        & Object+Style               & 0.268 & 0.288 & 0.357 & 0.242 & 0.156 & 0.167 & 0.214 & 0.142 & 0.118 & 0.147 & 0.231 & 0.128 & 0.174 & 0.213 & 0.314 & 0.193 \\ \bottomrule
\end{tabular}}
\caption{Evaluation for in-domain test cases. Along with Table~\ref{tab:outdomain}, these evaluation are visualized in Figure~\ref{fig:bar_plot},\ref{fig:outdomain},\ref{fig:ablation},\ref{fig:sup_bar_plot}.}
\label{tab:indomain}
\end{table*}

\begin{table*}[]
\resizebox{\linewidth}{!}{
\begin{tabular}{llcccccccccccccccc}
\toprule
\multicolumn{2}{l}{Source}                           & \multicolumn{8}{c}{ImageNet-Unseen}                           & \multicolumn{8}{c}{Artchive}                                  \\  \cmidrule(lr){1-2}\cmidrule(lr){3-10} \cmidrule(lr){11-18}
\multicolumn{2}{l}{Prompts}                          & \multicolumn{4}{c}{GPT}       & \multicolumn{4}{c}{Medium}    & \multicolumn{4}{c}{GPT}       & \multicolumn{4}{c}{Object}    \\ \cmidrule(lr){1-2}\cmidrule(lr){3-6}\cmidrule(lr){7-10} \cmidrule(lr){11-14}\cmidrule(lr){15-18}
$F_\text{base}$         & Method                     & R@5   & R@10  & R@100 & mAP   & R@5   & R@10  & R@100 & mAP   & R@5   & R@10  & R@100 & mAP   & R@5   & R@10  & R@100 & mAP   \\ \midrule
\multirow{8}{*}{CLIP}   & Pretrained                 & 0.580 & 0.644 & 0.818 & 0.500 & 0.239 & 0.285 & 0.472 & 0.201 & 0.186 & 0.234 & 0.440 & 0.198 & 0.140 & 0.175 & 0.335 & 0.144 \\ \cdashline{2-18} 
                        & Object                     & 0.710 & 0.754 & 0.868 & 0.645 & 0.511 & 0.567 & 0.733 & 0.447 & 0.249 & 0.305 & 0.512 & 0.282 & 0.188 & 0.231 & 0.396 & 0.207 \\
                        & Style                      & 0.664 & 0.716 & 0.853 & 0.587 & 0.327 & 0.378 & 0.567 & 0.279 & 0.264 & 0.325 & 0.565 & 0.307 & 0.203 & 0.251 & 0.447 & 0.228 \\ \cdashline{2-18} 
                        & Object+Style (No reg.)     & 0.524 & 0.588 & 0.777 & 0.438 & 0.275 & 0.327 & 0.517 & 0.220 & 0.057 & 0.078 & 0.185 & 0.059 & 0.042 & 0.059 & 0.145 & 0.043 \\ \cdashline{2-18} 
                        & Object+Style (Channel)     & 0.593 & 0.653 & 0.818 & 0.510 & 0.256 & 0.303 & 0.487 & 0.216 & 0.207 & 0.259 & 0.486 & 0.226 & 0.159 & 0.201 & 0.382 & 0.169 \\
                        & Object+Style (MLP)         & 0.346 & 0.427 & 0.694 & 0.260 & 0.187 & 0.248 & 0.504 & 0.139 & 0.025 & 0.036 & 0.106 & 0.026 & 0.018 & 0.026 & 0.081 & 0.018 \\
                        & Object+Style (Strong Aug.) & 0.697 & 0.744 & 0.863 & 0.629 & 0.393 & 0.453 & 0.644 & 0.335 & 0.252 & 0.308 & 0.536 & 0.285 & 0.193 & 0.236 & 0.423 & 0.211 \\ \cdashline{2-18} 
                        & Object+Style               & 0.701 & 0.745 & 0.864 & 0.633 & 0.389 & 0.445 & 0.628 & 0.332 & 0.259 & 0.317 & 0.550 & 0.297 & 0.201 & 0.247 & 0.437 & 0.223 \\ \midrule
\multirow{8}{*}{DINO}   & Pretrained                 & 0.831 & 0.851 & 0.900 & 0.795 & 0.540 & 0.572 & 0.661 & 0.492 & 0.183 & 0.211 & 0.320 & 0.181 & 0.140 & 0.162 & 0.250 & 0.136 \\ \cdashline{2-18} 
                        & Object                     & 0.842 & 0.861 & 0.909 & 0.809 & 0.622 & 0.654 & 0.738 & 0.574 & 0.225 & 0.260 & 0.386 & 0.232 & 0.175 & 0.203 & 0.303 & 0.176 \\
                        & Style                      & 0.838 & 0.858 & 0.905 & 0.803 & 0.576 & 0.608 & 0.698 & 0.526 & 0.214 & 0.247 & 0.374 & 0.221 & 0.168 & 0.193 & 0.293 & 0.168 \\ \cdashline{2-18} 
                        & Object+Style (No reg.)     & 0.646 & 0.692 & 0.823 & 0.574 & 0.361 & 0.404 & 0.544 & 0.308 & 0.062 & 0.082 & 0.183 & 0.058 & 0.045 & 0.058 & 0.139 & 0.041 \\ \cdashline{2-18} 
                        & Object+Style (Channel)     & 0.784 & 0.809 & 0.877 & 0.741 & 0.437 & 0.472 & 0.572 & 0.391 & 0.136 & 0.162 & 0.272 & 0.132 & 0.104 & 0.123 & 0.201 & 0.097 \\
                        & Object+Style (MLP)         & 0.335 & 0.413 & 0.668 & 0.255 & 0.173 & 0.223 & 0.435 & 0.127 & 0.016 & 0.026 & 0.091 & 0.017 & 0.013 & 0.020 & 0.070 & 0.013 \\
                        & Object+Style (Strong Aug.) & 0.842 & 0.861 & 0.907 & 0.809 & 0.594 & 0.626 & 0.711 & 0.544 & 0.212 & 0.244 & 0.366 & 0.216 & 0.166 & 0.191 & 0.287 & 0.165 \\ \cdashline{2-18} 
                        & Object+Style               & 0.842 & 0.861 & 0.908 & 0.810 & 0.598 & 0.629 & 0.714 & 0.549 & 0.217 & 0.248 & 0.374 & 0.221 & 0.170 & 0.194 & 0.294 & 0.169 \\ \midrule
\multirow{5}{*}{MoCo}   & Pretrained                 & 0.761 & 0.786 & 0.852 & 0.717 & 0.460 & 0.493 & 0.586 & 0.408 & 0.169 & 0.198 & 0.314 & 0.164 & 0.131 & 0.152 & 0.246 & 0.125 \\ \cdashline{2-18} 
                        & Object                     & 0.792 & 0.813 & 0.874 & 0.753 & 0.542 & 0.573 & 0.658 & 0.490 & 0.215 & 0.250 & 0.387 & 0.218 & 0.172 & 0.201 & 0.312 & 0.172 \\
                        & Style                      & 0.783 & 0.806 & 0.868 & 0.742 & 0.493 & 0.526 & 0.612 & 0.443 & 0.204 & 0.237 & 0.371 & 0.204 & 0.160 & 0.188 & 0.297 & 0.160 \\ \cdashline{2-18} 
                        & Object+Style (No reg.)     & 0.655 & 0.694 & 0.805 & 0.589 & 0.384 & 0.424 & 0.552 & 0.332 & 0.056 & 0.075 & 0.171 & 0.054 & 0.042 & 0.055 & 0.134 & 0.039 \\ \cdashline{2-18} 
                        & Object+Style               & 0.791 & 0.813 & 0.874 & 0.753 & 0.519 & 0.551 & 0.636 & 0.467 & 0.208 & 0.242 & 0.377 & 0.209 & 0.165 & 0.193 & 0.304 & 0.165 \\ \midrule
\multirow{5}{*}{ViT}    & Pretrained                 & 0.785 & 0.821 & 0.898 & 0.726 & 0.474 & 0.528 & 0.689 & 0.410 & 0.201 & 0.238 & 0.395 & 0.210 & 0.156 & 0.184 & 0.309 & 0.161 \\ \cdashline{2-18} 
                        & Object                     & 0.818 & 0.844 & 0.908 & 0.773 & 0.567 & 0.615 & 0.749 & 0.505 & 0.192 & 0.226 & 0.368 & 0.200 & 0.146 & 0.171 & 0.278 & 0.148 \\
                        & Style                      & 0.785 & 0.820 & 0.898 & 0.727 & 0.479 & 0.534 & 0.700 & 0.415 & 0.208 & 0.245 & 0.400 & 0.218 & 0.159 & 0.187 & 0.310 & 0.164 \\ \cdashline{2-18} 
                        & Object+Style (No reg.)     & 0.515 & 0.585 & 0.784 & 0.423 & 0.297 & 0.353 & 0.541 & 0.238 & 0.061 & 0.082 & 0.184 & 0.058 & 0.046 & 0.061 & 0.148 & 0.042 \\ \cdashline{2-18} 
                        & Object+Style               & 0.820 & 0.847 & 0.912 & 0.772 & 0.539 & 0.590 & 0.737 & 0.474 & 0.206 & 0.244 & 0.398 & 0.216 & 0.159 & 0.187 & 0.308 & 0.163 \\ \midrule
\multirow{5}{*}{ALADIN} & Pretrained                 & 0.348 & 0.388 & 0.530 & 0.298 & 0.133 & 0.150 & 0.210 & 0.117 & 0.130 & 0.167 & 0.332 & 0.133 & 0.119 & 0.153 & 0.321 & 0.124 \\ \cdashline{2-18} 
                        & Object                     & 0.471 & 0.506 & 0.621 & 0.427 & 0.202 & 0.220 & 0.295 & 0.182 & 0.206 & 0.250 & 0.425 & 0.219 & 0.185 & 0.228 & 0.396 & 0.200 \\
                        & Style                      & 0.342 & 0.380 & 0.515 & 0.297 & 0.145 & 0.161 & 0.220 & 0.128 & 0.149 & 0.183 & 0.343 & 0.151 & 0.137 & 0.173 & 0.335 & 0.145 \\ \cdashline{2-18} 
                        & Object+Style (No reg.)     & 0.264 & 0.317 & 0.510 & 0.208 & 0.115 & 0.140 & 0.237 & 0.092 & 0.063 & 0.086 & 0.201 & 0.059 & 0.060 & 0.081 & 0.190 & 0.057 \\ \cdashline{2-18} 
                        & Object+Style               & 0.443 & 0.481 & 0.604 & 0.396 & 0.172 & 0.186 & 0.249 & 0.155 & 0.167 & 0.208 & 0.373 & 0.174 & 0.155 & 0.194 & 0.362 & 0.165 \\ \midrule
\multirow{5}{*}{SSCD}   & Pretrained                 & 0.601 & 0.627 & 0.698 & 0.566 & 0.264 & 0.281 & 0.342 & 0.245 & 0.151 & 0.171 & 0.241 & 0.142 & 0.123 & 0.140 & 0.203 & 0.115 \\ \cdashline{2-18} 
                        & Object                     & 0.622 & 0.644 & 0.713 & 0.588 & 0.285 & 0.302 & 0.366 & 0.264 & 0.159 & 0.179 & 0.255 & 0.151 & 0.133 & 0.149 & 0.215 & 0.125 \\
                        & Style                      & 0.599 & 0.623 & 0.699 & 0.565 & 0.267 & 0.285 & 0.347 & 0.247 & 0.154 & 0.173 & 0.246 & 0.144 & 0.126 & 0.142 & 0.207 & 0.118 \\ \cdashline{2-18} 
                        & Object+Style (No reg.)     & 0.141 & 0.171 & 0.296 & 0.113 & 0.055 & 0.065 & 0.113 & 0.044 & 0.020 & 0.026 & 0.071 & 0.017 & 0.018 & 0.024 & 0.063 & 0.015 \\ \cdashline{2-18} 
                        & Object+Style               & 0.621 & 0.644 & 0.714 & 0.586 & 0.282 & 0.298 & 0.362 & 0.261 & 0.159 & 0.179 & 0.254 & 0.150 & 0.131 & 0.148 & 0.213 & 0.123 \\ \bottomrule
\end{tabular}}
\caption{Evaluation for out-of-domain test cases. Along with Table~\ref{tab:indomain}, these evaluation are visualized in Figure~\ref{fig:bar_plot},\ref{fig:outdomain},\ref{fig:ablation},\ref{fig:sup_bar_plot}.}
\label{tab:outdomain}
\end{table*}

\section{Change log.}
\myparagraph{v1} Initial release.

\myparagraph{v2} ICCV 2023 camera ready. Updated Figure~\ref{fig:sd_nn} and Figure~\ref{fig:supp_sd_nn} for attribution on the full LAION-400M dataset. Corrected errors in Figure~\ref{fig:softmax}. Minor changes in text.

\end{document}